\documentclass[11pt]{article}

\usepackage[final]{acl}

\usepackage{times}
\usepackage{latexsym}
\usepackage[T1]{fontenc}
\usepackage[utf8]{inputenc}
\usepackage{microtype}
\usepackage{inconsolata}

\usepackage{booktabs}
\usepackage{subcaption}
\usepackage{dblfloatfix}
\usepackage{float}
\usepackage{makecell}
\usepackage{amsmath}
\usepackage{mathtools}
\usepackage{amsthm}
\usepackage{graphicx}
\usepackage{enumitem}
\usepackage[most]{tcolorbox}
\usepackage{xurl}
\usepackage{fvextra}
\usepackage{amsmath,amssymb,amsfonts}
\fvset{
  breaklines=true,
  breakanywhere=true,
  breaksymbolleft={},
  breaksymbolright={},
  fontsize=\small
}

\graphicspath{{image_pdf/}}

\newif\iffullappendix
\fullappendixtrue

\newcommand{\n}[1]{#1}

\tcbset{
  promptboxstyle/.style={
    breakable,
    enhanced,
    colback=white,
    colframe=gray!70,
    fonttitle=\bfseries,
    coltitle=white,
    sharp corners,
    boxrule=0.8pt,
    arc=2pt,
    left=1mm,
    right=1mm,
    top=0.7mm,
    bottom=0.7mm,
  }
}

\theoremstyle{definition}

\theoremstyle{remark}

\title{MMLDSum-LLM: Multimodal Long-Document Summarization with Visual-Alignment and Keyword-Aware}

\author{
Xianpeng Zhang \and Jiahua Yang \and Dongyu Chen \and Lei Zhang \and Jian Ma \\
Xu Guohuan \and Haonan Lu \and Tianhuang Su \and Chuangchuang Wang \and Kai Tang \\
OPPO Guangdong Mobile Telecommunications Co., Ltd.
}

\begin{document}
\maketitle

\begin{abstract}
Multimodal long documents are core carriers of professional knowledge, where critical evidence is sparsely distributed across paragraphs and modalities. This easily causes key information omission and cross-modal hallucinations in summarization by multimodal LLMs. These issues stem from attention drift in long-range dependency modeling and gaps in inter-modal alignment. To address this, we introduce \textbf{MMLDSum-Bench}, a high-quality benchmark for multimodal long-document summarization, covering multiple domains, context-length scales, and visual-textual modality distributions. We further propose \textbf{MMLDSum-LLM}, a reproducible two-stage training framework that combines supervised fine-tuning with \emph{visual-alignment weighted loss} and \emph{keyword-aware weighted loss}, followed by GRPO with a multi-objective reward (keyword coverage, image-text alignment, ROUGE, and length control). Extensive experiments on MMLDSum-Bench evaluate our approach against leading closed-source and open-source multimodal models under a unified protocol that incorporates LLM-as-a-judge scoring, atomic-claim precision/recall, image-text alignment (ITA), and ROUGE. The results demonstrate that our approach significantly improves key-information coverage and cross-modal consistency.
\end{abstract}

\section{Introduction}
In an era of information explosion~\cite{goyal2022news}, multimodal long documents, such as academic papers, medical reports, and financial annual reports, have become the dominant medium for knowledge transmission in professional domains. Such documents integrate multiple modalities, including text, figures, and tables, with each contributing distinct yet complementary information. Crucially, these modalities do not function in isolation but mutually reinforce and corroborate one another, collectively delivering the full informational content of the document. Multimodal long-document summarization aims to condense such a document into a concise, coherent natural-language summary that faithfully captures the salient information across all modalities, preserves cross-modal evidential consistency, and retains the key factual relations between textual arguments and their supporting visual elements.

\begin{figure}[t]
  \centering
  \includegraphics[width=0.9\columnwidth]{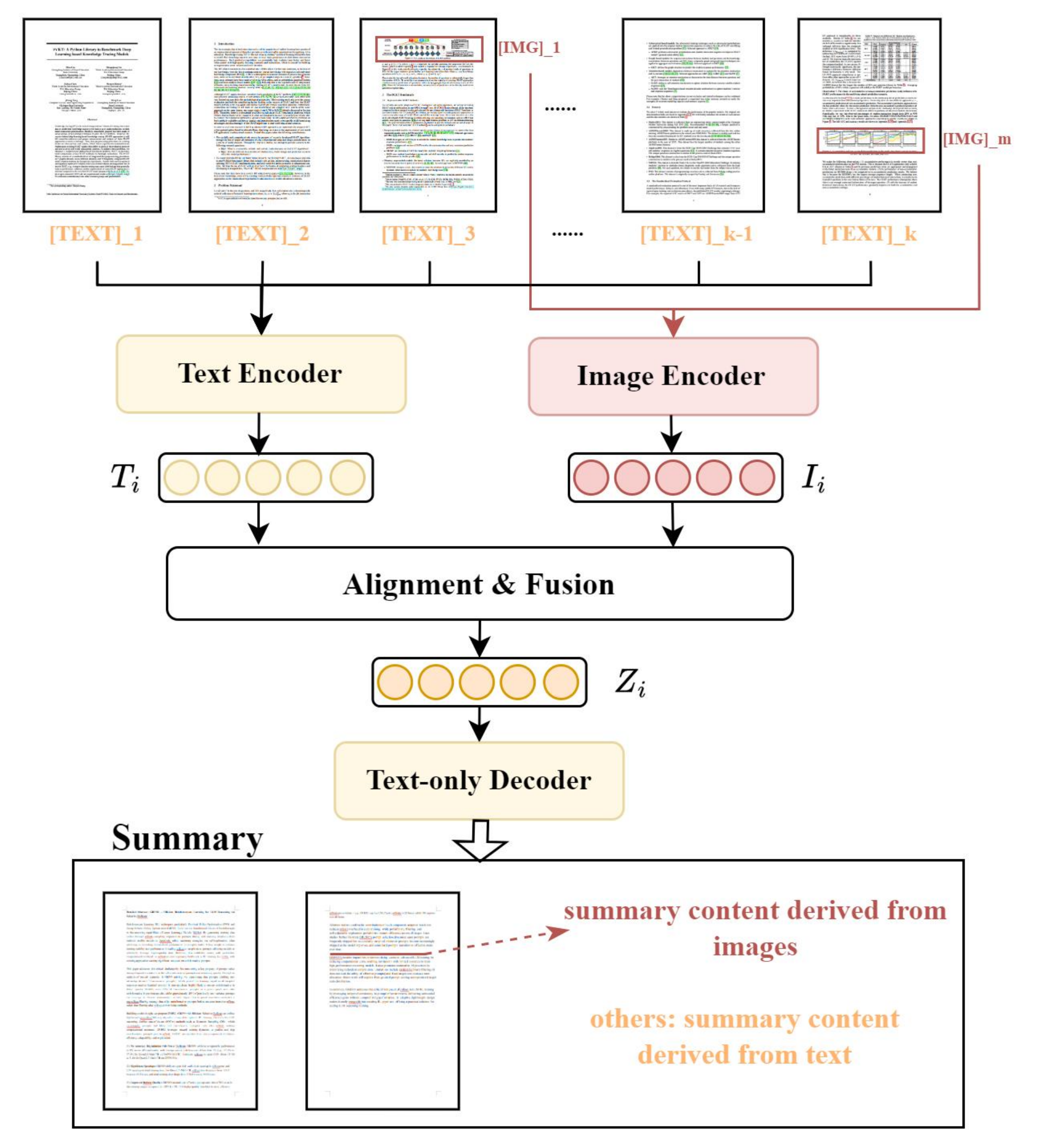}
  \caption{Illustration of conventional multimodal long-document summarization.}
  \label{fig:overview}
\end{figure}

Early summarization studies focused on extractive methods (e.g., TF-IDF, TextRank) and later shifted to neural abstractive models~\cite{rush2015neural}. With the rise of large language models, summarization has benefited from stronger generation quality and controllable prompting. In parallel, multimodal summarization extends beyond text by incorporating images and other modalities, often requiring explicit cross-modal alignment to avoid modality suppression and hallucinations~\cite{jangra2023survey}. However, most existing multimodal summarization benchmarks and methods primarily focus on short contexts or domain-specific settings (e.g., dialogue/video), and do not capture the sparse, cross-modally dispersed distribution of key evidence in multimodal long documents~\cite{kumbhar2023current,khilji2023multimodal}.

At present, multimodal long-document summarization faces severe challenges at both the data and methodological levels~\cite{koh2022empirical}. \textbf{On the data side}, most existing multimodal summarization datasets are confined to specific domains — such as dialogues, news, and clinical reports — or limited to short contexts, leaving long multimodal documents with diverse visual-textual modality distributions substantially underrepresented. Although several long-context multimodal benchmarks have recently emerged, summarization-specific supervision and evaluation protocols under long-context settings remain scarce~\cite{wang2025mmlongbench}. \textbf{On the method side}, models are required to jointly address long-range dependency modeling, cross-modal grounding, and information selection. Failures in these aspects typically manifest as both missing key evidence and cross-modal hallucinations. As illustrated in Figure~\ref{fig:overview}, conventional multimodal long-document summarization approaches generally adopt a pipeline architecture consisting of modality-specific encoding, feature alignment and fusion, and decoder-based generation. Textual and visual features are first extracted independently by text and image encoders, aligned and fused into a joint representation, and subsequently decoded to produce a textual summary. Some studies further select images that are most semantically relevant to the generated summary to yield multimodal summary outputs. However, such paradigms remain fundamentally limited by constrained long-sequence modeling capacity, unstable cross-modal semantic alignment, and inadequate mechanisms for effective information selection~\cite{hua2025v2xum}.

These failures stem from three intertwined challenges: (i)~attention drift over long sequences causes models to over-attend to local context and miss globally salient evidence~\cite{ouyang2022training}; (ii)~cross-modal misalignment causes visually distant evidence to be suppressed or hallucinated; and (iii)~standard SFT objectives treat all tokens uniformly, giving insufficient weight to sparse but critical evidence spans.

To address these challenges, we propose MMLDSum-LLM (Multimodal Long-Document Summarization with Visual-Alignment and Keyword-Aware Training), a two-stage framework integrating supervised fine-tuning (SFT) and group relative policy optimization (GRPO)~\cite{shao2024deepseekmath}. 
Motivated by a cognitive anchoring strategy, in which readers first anchor core concepts and salient visuals before organizing supporting details, we design a composite weighted SFT loss with two complementary components: a visual-alignment weight that amplifies learning on image-associated spans, and a keyword-aware weight that emphasizes TF-IDF-filtered key entities.
GRPO then optimizes sequence-level objectives via multi-objective verifiable rewards for keyword coverage, image-text alignment, ROUGE, and length control.
We also introduce MMLDSum-Bench, a benchmark covering six domains, five context-length scales (4k--64k tokens), and four visual-textual modality distribution categories, providing a comprehensive testbed for this task.

\textbf{Our contributions are summarized as follows:}
\begin{itemize}[leftmargin=*]
\sloppy

\item We construct \textbf{MMLDSum-Bench}, a high-quality benchmark for multimodal long-document summarization, providing a comprehensive and realistic testbed for this task.

\item We design a systematic \textbf{evaluation protocol} encompassing LLM-as-a-judge scoring, atomic-claim precision/recall/F1, image-text alignment (ITA), and ROUGE, and conduct a unified comparative evaluation of state-of-the-art closed-source and open-source multimodal models on MMLDSum-Bench.

\item We propose \textbf{MMLDSum-LLM}, a two-stage training framework that combines visual-alignment and keyword-aware weighted SFT with GRPO-based reinforcement learning using multi-objective verifiable rewards. Experiments demonstrate that MMLDSum-LLM significantly improves key-information coverage and cross-modal consistency.
\end{itemize}

\section{Related Work}
\subsection{Multimodal Summarization}
LLMs have substantially improved text summarization in generation quality and instruction following~\cite{narayan2021planning,adams2023sparse}. Multimodal summarization extends this by incorporating images and other modalities via modality-specific encoders, cross-modal fusion, and contrastive alignment~\cite{li2018multi,he2023align}, with retrieval-augmented methods further improving visual grounding~\cite{rafi2024sct}. Research spans domain-specific settings including medical imaging~\cite{ghosh2024sights,lu2022research} and dynamic scenarios such as dialogue and video summarization~\cite{lu2024modality,qiu2024mmsum,hua2025v2xum,yang2024multi,mahon2024modular,tan2025enhancing}.

\subsection{Long-Context Vision--Language Models}
Long-Context VLMs (LCVLMs)~\cite{song2025bridge} enable end-to-end multimodal understanding at scale~\cite{wang2025mmlongbench}, but remain limited for long-document summarization: their alignment modules are designed for shorter sequences, causing semantic drift when evidence is asynchronously distributed across long documents~\cite{bai2024longwriter,wan2025qwenlong}, and pre-training objectives target general understanding rather than the selective compression required for quality summaries~\cite{deng2025longdocurl}. MMLDSum-LLM directly addresses these gaps through explicit visual-alignment weighting and keyword-aware supervised training.

\section{MMLDSum-Bench}

The MMLDSum-Bench benchmark targets the \textbf{multimodal long-document summarization} task: given the textual content of a document and its associated image set, the model is required to generate a natural-language summary under a length constraint that captures core factual information and critical visual evidence while preserving cross-modal consistency. The benchmark contains approximately 5k ($5{,}149$) multimodal long documents paired with over 40k associated images across diverse domains. We stratify documents into five context-length scales ($4$k--$64$k tokens, with an average length of $\sim25$k tokens) and categorize the data into four categories of visual-textual modality distributions, spanning the full spectrum from heavily text-dominant to heavily image-dominant settings.

\begin{figure}[t]
  \centering
  \includegraphics[width=1.0\columnwidth]{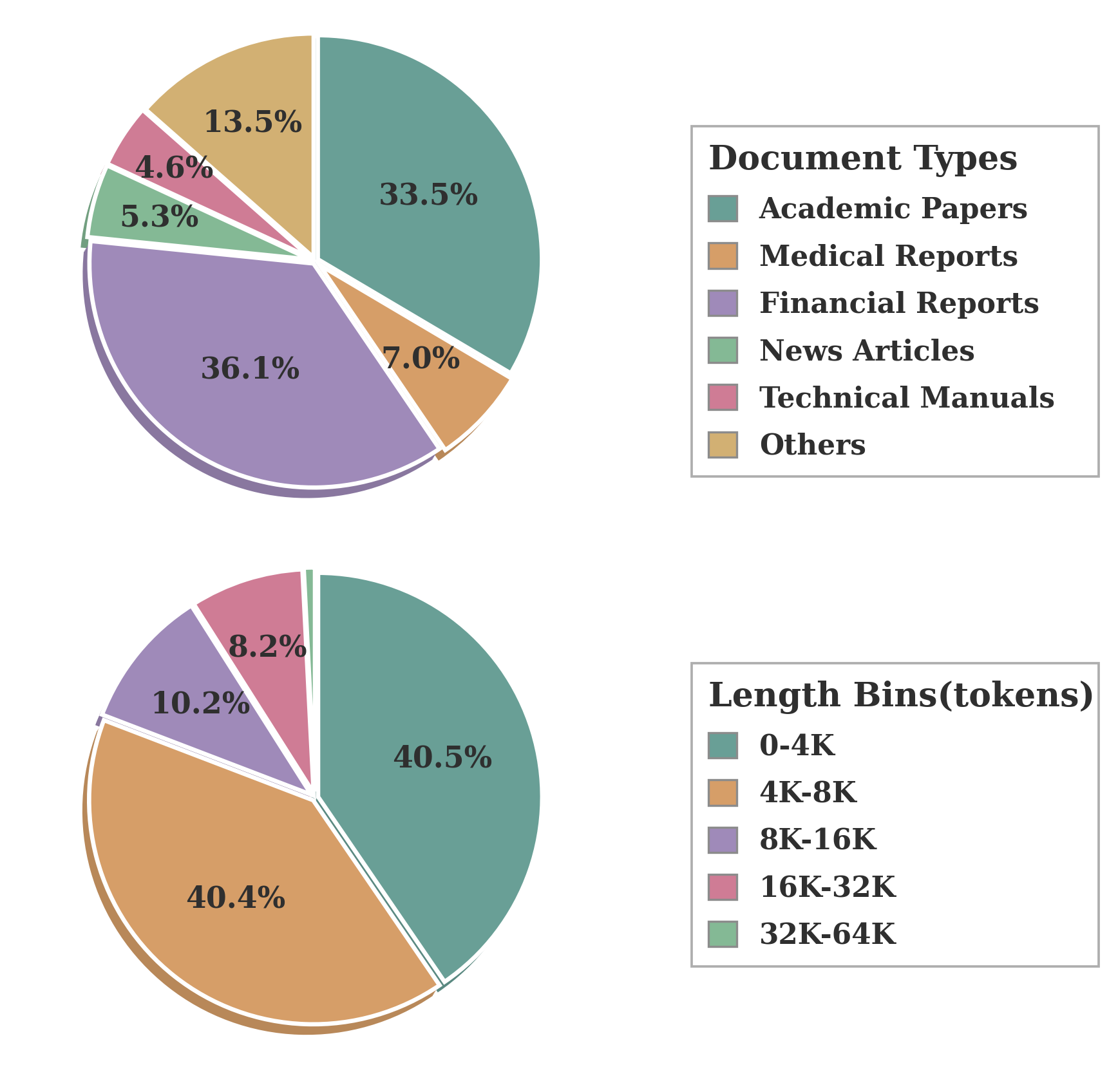}
  \caption{Domain distribution of MMLDSum-Bench.}
  \label{fig:datasets}
\end{figure}

As illustrated in Figure~\ref{fig:datasets}, the corpus covers multiple domains, including academia, medicine, finance, news, technology and others, enabling representative sampling of both narrative-heavy and evidence-heavy documents. In terms of context length, the dataset is concentrated in the $4$k--$16$k range while also containing a substantial number of samples in the $16$k--$64$k regime, which is sufficient for evaluating long-context summarization performance of multimodal large models. Table~\ref{tab:datasets} further shows broad coverage across modality distributions. Although the benchmark is dominated by heavily text-dominant documents, accounting for $80.9\%$ ($4{,}163$ samples), it also includes meaningful proportions of lightly text-dominant ($11.9\%$, $611$ samples) and lightly image-dominant ($6.6\%$, $342$ samples) documents.
These distribution characteristics ensure that the benchmark provides comprehensive coverage across three key dimensions (\textbf{domain, context length, and visual-textual modality distribution}) rather than a single narrow regime, thereby establishing a realistic and reliable data environment for multimodal long-document summarization research.

\begin{table}[t]
\centering
\setlength{\tabcolsep}{0.55pt}
\small
\caption{Distribution of Image--Text Ratios in the Dataset}
\label{tab:datasets}
\begin{tabular}{l|c|c|c}
\toprule
Type & Image Ratio & Count & Percentage (\%) \\
\midrule
Heavily Text-Dominant & 0--0.25 & 4163 & 80.9 \\
Lightly Text-Dominant & 0.25--0.5 & 611 & 11.9 \\
Lightly Image-Dominant & 0.5--0.75 & 342 & 6.6 \\
Heavily Image-Dominant & 0.75--1 & 33 & 0.6 \\
\bottomrule
\end{tabular}
\end{table}

As shown in Figure~\ref{fig:framework_data}, we employ a three-stage pipeline to balance quality and cost:

(i) \textbf{Data Processing}: This stage performs document chunking and global signal extraction. Each document is segmented by paragraph boundaries under a length threshold (approximately 3k tokens), with images assigned to chunks according to their original positions or adjacent paragraphs. Doubao-1.5-pro-256k is then used to extract global signals, including topics, outlines, and key entities.

(ii) \textbf{Multimodal Summary Generation}: Gemini-2.5-Pro first produces local summaries for individual chunks. These local summaries are then fused with the extracted global signals (topics, outlines, key entities) to generate candidate global summaries.

(iii) \textbf{Quality Verification and Regeneration}: Candidate summaries are evaluated through a multi-model scoring-and-voting mechanism (GPT-4o, Doubao-seed-1.6, Gemini-2.5-Pro) across five dimensions: completeness, accuracy, coherence, conciseness, and overall quality. A candidate is accepted only when all three models assign scores above a predefined threshold; otherwise, the summary generation process is re-executed until the consensus criterion is met.

To ensure robust dataset quality assessment, we conduct stratified human evaluation on 600 summaries across domain, context-length scale, and visual-textual modality distribution. Each summary is evaluated along five dimensions (completeness, accuracy, coherence, conciseness, and overall quality), and we report per-dimension mean scores with 95\% confidence intervals, together with per-dimension inter-annotator agreement. We further perform claim-level manual verification on 200 atomic claims to directly assess factual correctness and evidence-grounding consistency. In addition, we quantify the contribution of the regeneration module by reporting before/after quality statistics for regenerated samples. Detailed protocols and full results are provided in Appendix~\ref{sec:appendix-quality-validation}. Overall, the evaluation indicates high annotation quality (overall mean score: 4.7/5.0; overall Cohen's Kappa: 0.83).

\begin{figure*}[t]
  \centering
  \includegraphics[width=\textwidth]{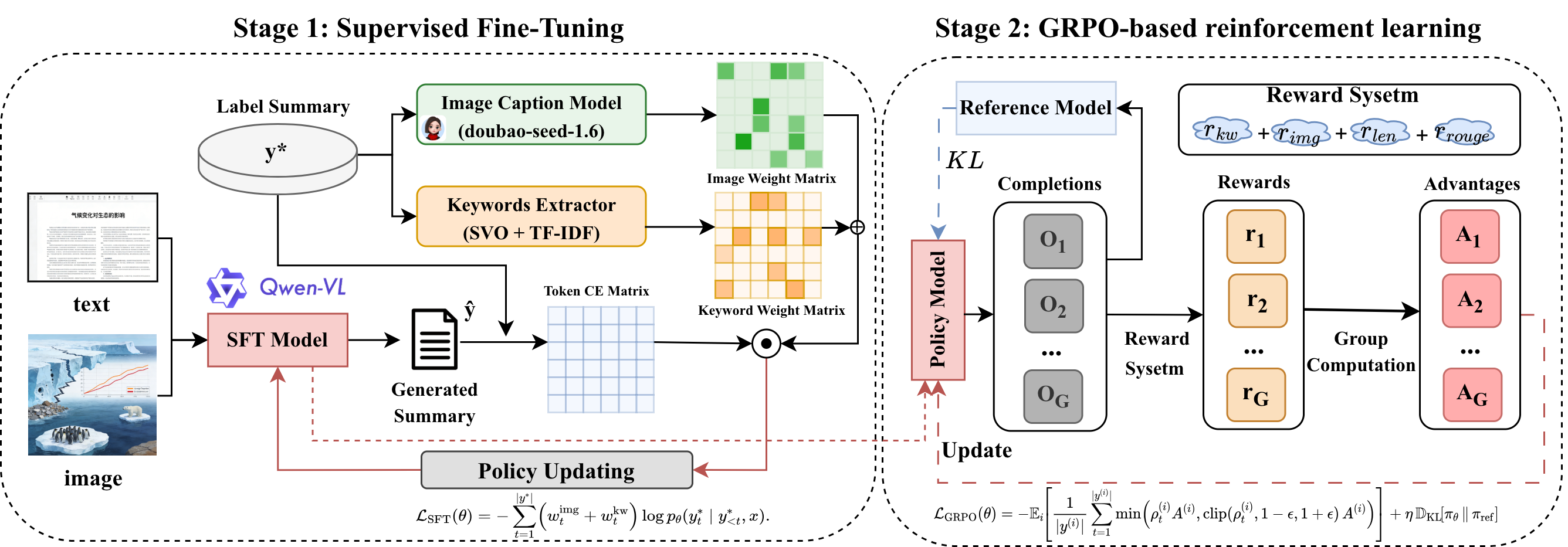}
  \caption{Overview of MMLDSum-LLM: alignment signal acquisition, weighted SFT, and GRPO-based reinforcement learning with multi-objective verifiable rewards.}
  \label{fig:framework_train}
\end{figure*}

\section{Methodology}
\subsection{Task Definition}
Let a multimodal document be \(x=(T,I)\), where \(T\) is the text token sequence and \(I\) is the image set.  
Given \(x\), the model generates a summary \(y=(y_1,\dots,y_n)\) with conditional distribution \(p_\theta(y\mid x)\).
As discussed in Section~1, multimodal long-document summarization mainly suffers from two issues: (i) omission of key information caused by attention drift over long contexts, and (ii) cross-modal hallucination caused by text--image misalignment. Therefore, our goal is not only to maximize conditional likelihood, but also to improve factual/visual evidence coverage and cross-modal consistency under a length budget.

As shown in Figure~\ref{fig:framework_train}, we optimize this goal with a two-stage framework.  
Stage 1 (anchor-weighted SFT) identifies textual and visual anchors and increases supervision on anchor-related spans to strengthen local grounding.  
Stage 2 (GRPO-based RL) optimizes sequence-level quality, including key-information coverage, cross-modal consistency, and conciseness.  
This local-to-global optimization forms the core of MMLDSum-LLM.

\subsection{Stage 1: Visual-Alignment and Keyword-Aware Weighted SFT}

\paragraph{Limitation of standard cross-entropy.}
Given training pairs $(x, y^\star)$, the standard token-level cross-entropy objective is:
\begin{equation}
  \mathcal{L}_{\mathrm{CE}}(\theta) = - \sum_{t=1}^{|y^\star|} \log p_\theta(y^\star_t \mid y^\star_{<t}, x).
  \label{eq:ce}
\end{equation}
This objective assigns equal importance to all reference tokens, which weakens supervision on sparse but critical evidence tokens. As a result, the model may miss key facts or generate visually unsupported content. We therefore introduce a weighted strategy to strengthen learning on evidence-critical positions.

Following the cognitive anchoring principle, we first amplify learning signals on visually grounded spans. During data construction, summary spans that describe or reference visual evidence are marked via special-token matching and regular-expression rules. We define an indicator $\mathbb{I}^{\mathrm{img}}_t \in \{0,1\}$ that equals 1 if token $y^\star_t$ belongs to a visually grounded span, and apply a per-token weight:
\begin{equation}
  w^{\mathrm{img}}_t = 1 + \lambda_{\mathrm{img}} \cdot \mathbb{I}^{\mathrm{img}}_t.
  \label{eq:wimg}
\end{equation}

In parallel, we build a keyword set $K$ as textual fact anchors. We extract subject--verb--object (SVO) tuples with a dependency parser, then apply TF-IDF filtering to keep domain-salient entities and relations. Let $\mathbb{I}^{\mathrm{kw}}_t \in \{0,1\}$ indicate whether token $y^\star_t$ matches an extracted keyword:
\begin{equation}
  w^{\mathrm{kw}}_t = 1 + \lambda_{\mathrm{kw}} \cdot \mathbb{I}^{\mathrm{kw}}_t.
  \label{eq:wkw}
\end{equation}

The final SFT loss fuses both weights additively to amplify learning signals on visual evidence and key facts:
\begin{equation}
  \mathcal{L}_{\mathrm{SFT}}(\theta) = - \sum_{t=1}^{|y^\star|} \left(w^{\mathrm{img}}_t + w^{\mathrm{kw}}_t\right)\, \log p_\theta(y^\star_t \mid y^\star_{<t}, x).
  \label{eq:sft}
\end{equation}

We use additive fusion so each signal contributes independently: tokens matched by either type are still reinforced, unlike multiplicative fusion, which mainly boosts rare co-occurrences. Hyperparameters $\lambda_{\mathrm{img}}$ and $\lambda_{\mathrm{kw}}$ control weighting strength; values in the range 5--7 provide a good balance between evidence coverage and fluency.

\subsection{Stage 2: GRPO-Based Reinforcement Learning}

Visual-alignment and keyword-aware weighted SFT strengthens token-level supervision on key evidence, but it is still imitation learning and remains tied to the training distribution. It also cannot directly optimize summary-level properties—key-information coverage, cross-modal consistency, and conciseness. To address this, we add a second stage using GRPO~\cite{shao2024deepseekmath}, which evaluates each sample against the within-group mean of $G$ candidate summaries. We use a composite reward with four components:
\begin{equation}
  r(y;x) = \alpha\, r_{\mathrm{kw}} + \beta\, r_{\mathrm{img}} + \gamma\, r_{\mathrm{rouge}} + \delta\, r_{\mathrm{len}}.
  \label{eq:reward}
\end{equation}

\begin{itemize}[leftmargin=*]
\item $r_{\mathrm{kw}}$ (\textit{keyword coverage}): mitigates key-information omission by measuring precision, recall, and F1 between generated-summary keywords and source fact anchors.
\item $r_{\mathrm{img}}$ (\textit{image-text alignment}): mitigates cross-modal hallucination by computing semantic similarity between summary segments and image captions from an auxiliary captioning model.
\item $r_{\mathrm{rouge}}$ (\textit{ROUGE score}): uses the average of ROUGE-1/2/L against the reference summary as a general quality signal.
\item $r_{\mathrm{len}}$ (\textit{length control}): discourages overly long outputs and controls RL-induced length inflation.
\end{itemize}

All four rewards are rule-based and deterministic, casting training as reinforcement learning with verifiable rewards (RLVR) and avoiding costly, unstable LLM-based reward models. We set $\alpha = 0.5$, $\beta = 0.2$, $\gamma = 0.15$, and $\delta = 0.15$, prioritizing keyword coverage because key-information omission is the dominant failure mode in preliminary experiments.

For each input $x$, we sample a group of $G$ candidate summaries
$\{y^{(i)}\}_{i=1}^{G} \sim \pi_{\theta_{\mathrm{old}}}(\cdot \mid x)$
from the policy snapshot and score each one with the composite reward
$r^{(i)}$ defined in Eq.~\ref{eq:reward}. Following GRPO~\cite{shao2024deepseekmath},
we standardize rewards within the group:
\begin{equation}
  A^{(i)} = \frac{r^{(i)} - \mathrm{mean}\!\left(\{r^{(j)}\}_{j=1}^{G}\right)}
                 {\mathrm{std}\!\left(\{r^{(j)}\}_{j=1}^{G}\right) + \varepsilon},
  \label{eq:adv}
\end{equation}

The policy is then updated with a token-level clipped objective regularized
toward a fixed reference policy $\pi_{\mathrm{ref}}$, which we initialize
from the Stage~1 SFT checkpoint and keep frozen throughout RL:
\begin{equation}
\begin{aligned}
  \mathcal{L}_{\mathrm{GRPO}}(\theta)
  ={}& - \mathbb{E}_{i}\!\left[\frac{1}{|y^{(i)}|}\sum_{t=1}^{|y^{(i)}|}
        \min\!\Big(\rho^{(i)}_t A^{(i)},\right. \\
     & \qquad \left. \mathrm{clip}(\rho^{(i)}_t,\, 1{-}\epsilon,\, 1{+}\epsilon)\, A^{(i)}\Big)\right] \\
     & + \eta\, \mathbb{D}_{\mathrm{KL}}\!\left[\pi_\theta \,\|\, \pi_{\mathrm{ref}}\right],
\end{aligned}
  \label{eq:grpo}
\end{equation}
where the per-token importance ratio is
\begin{equation}
  \rho^{(i)}_t = \frac{\pi_\theta(y^{(i)}_t \mid y^{(i)}_{<t}, x)}
                       {\pi_{\theta_{\mathrm{old}}}(y^{(i)}_t \mid y^{(i)}_{<t}, x)},
  \label{eq:ratio}
\end{equation}  
The KL term preserves Stage~1 priors (visual alignment and keyword grounding) while allowing stable optimization of summary-level rewards.

The two stages are complementary: Stage~1 improves local evidence grounding through token weighting, and Stage~2 improves global summary quality and generalization through reward-driven exploration.

\begin{table*}[!t]
    \caption{Comparison results on MMLDSum-Bench across closed-source models, open-source models, other methods, and our MMLDSum-LLM variants. Bold numbers denote the best-performing metrics.}
  \label{tab:main}
  \centering
  \setlength{\tabcolsep}{1.0pt}
  \renewcommand{\arraystretch}{0.85}
  \footnotesize
  \resizebox{\textwidth}{!}{%
  \begin{tabular}{l c ccccc ccccc cc c ccc}
    \toprule
    \multicolumn{1}{c}{Model} & \multicolumn{1}{c}{\makecell{Max \\ ctx}}
 &
    \multicolumn{5}{c}{GPT-4o score} &
    \multicolumn{5}{c}{GPT-5 score} &
    \multicolumn{2}{c}{Atomic claim} &
    \multicolumn{1}{c}{ITA} &
    \multicolumn{3}{c}{ROUGE} \\
    \cmidrule(lr){3-7} \cmidrule(lr){8-12} \cmidrule(lr){13-14} \cmidrule(lr){15-15} \cmidrule(lr){16-18}
    & & Comp. & Acc. & Conc. & Coh. & Overall & Comp. & Acc. & Conc. & Coh. & Overall & R & F1 & ITA-R & R-1 & R-2 & R-L \\
    \midrule
    \multicolumn{18}{l}{\textit{Closed-source models}} \\
    step-1o-vision-32k & 32k & \n{4.37} & \n{4.88} & \n{4.95} & \n{4.94} & \n{4.57} & \n{3.19} & \textbf{\n{4.28}} & \n{4.64} & \n{4.82} & \n{3.51} & \n{0.47} & \n{0.60} & \n{0.49} & \n{0.42} & \n{0.20} & \n{0.25} \\
    claude-4-sonnet & 1000k & \n{4.34} & \n{4.91} & \n{4.94} & \n{4.95} & \n{4.52} & \n{3.48} & \n{3.79} & \textbf{\n{4.67}} & \n{4.60} & \n{3.58} & \n{0.67} & \n{0.75} & \n{0.59} & \n{0.50} & \n{0.24} & \n{0.30} \\
    qwen-vl-max & 128k & \n{4.57} & \n{4.94} & \n{4.95} & \n{4.98} & \n{4.75} & \n{4.00} & \n{3.64} & \n{4.16} & \n{4.85} & \n{3.70} & \n{0.66} & \n{0.73} & \textbf{\n{0.72}} & \n{0.52} & \n{0.23} & \n{0.29} \\
    qwen3-vl-plus & 256k & \textbf{\n{4.66}} & \textbf{\n{4.96}} & \textbf{\n{4.96}} & \textbf{\n{4.99}} & \textbf{\n{4.83}} & \textbf{\n{4.08}} & \n{3.79} & \n{4.07} & \n{4.88} & \n{3.76} & \n{0.71} & \n{0.77} & \n{0.71} & \textbf{\n{0.55}} & \n{0.24} & \n{0.30} \\
    doubao-seed-1.6 & 256k & \n{4.58} & \n{4.94} & \textbf{\n{4.96}} & \n{4.98} & \n{4.77} & \n{3.98} & \n{4.06} & \n{4.49} & \textbf{\n{4.90}} & \textbf{\n{3.87}} & \n{0.71} & \n{0.77} & \n{0.66} & \textbf{\n{0.55}} & \textbf{\n{0.27}} & \textbf{\n{0.34}} \\
    gpt-5 & 128k & \n{4.64} & \n{4.94} & \n{4.86} & \n{4.96} & \n{4.79} & -- & -- & -- & -- & -- & \textbf{\n{0.90}} & \textbf{\n{0.85}} & \textbf{\n{0.72}} & \n{0.53} & \n{0.20} & \n{0.30} \\
    \midrule
    \multicolumn{18}{l}{\textit{Open-source models}} \\
    phi-4-multimodal-instruct & 128k & \n{1.87} & \n{1.82} & \n{2.28} & \n{2.10} & \n{1.76} & \n{1.25} & \n{1.31} & \n{1.56} & \n{1.82} & \n{1.28} & \n{0.16} & \n{0.18} & \n{0.29} & \n{0.09} & \n{0.02} & \n{0.06} \\
    qwen2.5-vl-32b-instruct & 128k & \n{3.78} & \n{4.51} & \n{4.33} & \n{4.61} & \n{4.04} & \n{2.30} & \n{2.45} & \n{2.47} & \n{3.34} & \n{2.37} & \n{0.42} & \n{0.52} & \n{0.63} & \n{0.33} & \n{0.10} & \n{0.16} \\
    qwen2.5-vl-72b-instruct & 128k & \n{3.67} & \n{4.42} & \n{4.11} & \n{4.46} & \n{3.90} & \n{2.24} & \n{2.43} & \n{2.32} & \n{3.09} & \n{2.24} & \n{0.43} & \n{0.51} & \n{0.65} & \n{0.25} & \n{0.07} & \n{0.12} \\
    internvl3.5-14b-instruct & 32k & \n{4.11} & \n{4.76} & \n{4.81} & \n{4.83} & \n{4.36} & \n{2.86} & \n{3.53} & \n{4.36} & \n{4.51} & \n{3.11} & \n{0.47} & \n{0.58} & \n{0.54} & \n{0.35} & \n{0.15} & \n{0.20} \\
    internvl3.5-38b-instruct & 32k & \n{4.01} & \n{4.74} & \n{4.76} & \n{4.80} & \n{4.29} & \n{2.81} & \n{3.63} & \n{4.34} & \n{4.45} & \n{3.12} & \n{0.42} & \n{0.54} & \n{0.51} & \n{0.30} & \n{0.12} & \n{0.17} \\
    gemma3-12b & 128k & \n{4.20} & \n{4.81} & \n{4.89} & \n{4.90} & \n{4.43} & \n{2.96} & \n{3.39} & \n{4.49} & \n{4.56} & \n{3.17} & \n{0.46} & \n{0.58} & \n{0.59} & \n{0.36} & \n{0.16} & \n{0.21} \\
    gemma3-27b & 128k & \n{4.20} & \textbf{\n{4.86}} & \textbf{\n{4.93}} & \n{4.92} & \n{4.46} & \n{3.05} & \n{3.62} & \textbf{\n{4.61}} & \n{4.60} & \textbf{\n{3.31}} & \n{0.48} & \n{0.61} & \n{0.57} & \n{0.31} & \n{0.13} & \n{0.18} \\
    qwen3-vl-32b-instruct & 256k & \n{4.16} & \n{4.81} & \n{4.65} & \n{4.89} & \n{4.36} & \n{2.75} & \n{2.30} & \n{2.19} & \n{3.53} & \n{2.46} & \n{0.58} & \n{0.62} & \n{0.78} & \n{0.35} & \n{0.09} & \n{0.15} \\
    qwen3.5-vl-27b & 128k & \n{3.88} & \n{4.22} & \n{4.26} & \n{4.31} & \n{4.00} & \n{3.18} & \n{2.82} & \n{3.34} & \n{4.03} & \n{2.94} & \n{0.63} & \n{0.63} & \n{0.47} & \n{0.43} & \n{0.17} & \n{0.23} \\
    \midrule
    \multicolumn{18}{l}{\textit{Other methods}} \\
    qwen2.5-vl-7b-cod & 128k & \n{3.64} & \n{4.30} & \n{4.38} & \n{4.31} & \n{3.90} & \n{2.56} & \n{3.34} & \n{4.31} & \n{4.25} & \n{2.90} & \n{0.30} & \n{0.41} & \n{0.55} & \n{0.26} & \n{0.11} & \n{0.15} \\
    qwen3-vl-8b-cod & 256k & \n{4.20} & \n{4.71} & \n{4.56} & \n{4.73} & \n{4.45} & \n{3.71} & \n{2.83} & \n{3.99} & \n{4.63} & \n{3.25} & \n{0.63} & \n{0.69} & \n{0.60} & \n{0.37} & \n{0.14} & \n{0.21} \\
    longwriter-llama3.1-8b-caption & 128k & \n{3.58} & \n{4.31} & \n{4.48} & \n{4.34} & \n{3.87} & \n{2.42} & \n{3.76} & \n{4.47} & \n{4.39} & \n{2.88} & \n{0.29} & \n{0.40} & \n{0.41} & \n{0.18} & \n{0.07} & \n{0.11} \\
    longwriter-glm4-9b-caption & 128k & \n{3.42} & \n{4.30} & \n{4.36} & \n{4.25} & \n{3.72} & \n{2.53} & \textbf{\n{4.00}} & \n{3.58} & \n{3.88} & \n{2.80} & \n{0.37} & \n{0.44} & \n{0.60} & \n{0.24} & \n{0.08} & \n{0.12} \\
    \midrule
    \multicolumn{18}{l}{\textit{\textbf{Ours}}} \\
    qwen2.5-vl-3b-sft & 128k & \n{2.92} & \n{3.42} & \n{3.45} & \n{3.55} & \n{3.15} & \n{1.86} & \n{1.51} & \n{1.98} & \n{2.48} & \n{1.68} & \n{0.34} & \n{0.38} & \n{0.50} & \n{0.26} & \n{0.06} & \n{0.12} \\
    MMLDSum-qwen2.5-vl-3b & 128k & \n{3.48} & \n{4.18} & \n{4.08} & \n{4.37} & \n{3.79} & \n{2.26} & \n{1.82} & \n{2.43} & \n{3.10} & \n{2.00} & \n{0.53} & \n{0.54} & \n{0.76} & \n{0.38} & \n{0.12} & \n{0.17} \\
    qwen2.5-vl-7b-sft & 128k & \n{3.48} & \n{4.22} & \n{4.51} & \n{4.44} & \n{3.86} & \n{2.43} & \n{2.20} & \n{3.45} & \n{3.87} & \n{2.42} & \n{0.46} & \n{0.53} & \n{0.72} & \n{0.41} & \n{0.14} & \n{0.20} \\
    MMLDSum-qwen2.5-vl-7b & 128k & \n{3.82} & \n{4.65} & \n{4.60} & \n{4.79} & \n{4.13} & \n{2.63} & \n{2.47} & \n{3.66} & \n{4.01} & \n{2.58} & \n{0.54} & \n{0.59} & \n{0.87} & \n{0.51} & \n{0.21} & \n{0.26} \\
    qwen3-vl-8b-sft & 256k & \n{4.10} & \n{4.73} & \n{4.63} & \n{4.83} & \n{4.29} & \n{3.47} & \n{2.68} & \n{3.04} & \n{4.13} & \n{2.78} & \n{0.73} & \n{0.73} & \n{0.82} & \n{0.51} & \n{0.21} & \n{0.28} \\
    MMLDSum-qwen3-vl-8b & 256k & \textbf{\n{4.33}} & \n{4.85} & \n{4.78} & \textbf{\n{4.93}} & \textbf{\n{4.51}} & \textbf{\n{4.08}} & \n{3.76} & \n{3.65} & \textbf{\n{4.66}} & \n{3.21} & \textbf{\n{0.85}} & \textbf{\n{0.80}} & \textbf{\n{0.89}} & \textbf{\n{0.63}} & \textbf{\n{0.30}} & \textbf{\n{0.37}} \\
    \bottomrule
  \end{tabular}
  }
\end{table*}

\section{Experiments}
\subsection{Evaluation Metrics}
To comprehensively evaluate multimodal long-document summarization, we build a multidimensional automatic evaluation suite~\cite{liu2025mdseval,langston2024automated} with four complementary metric families. Each family focuses on a different quality dimension, and their combination enables cross-validation over semantic fidelity, cross-modal consistency, and surface-level text quality. If a model shows stable gains across all metrics, this provides strong evidence of substantive summary quality improvement.

\begin{itemize}[leftmargin=*]
\sloppy
\item \textbf{LLM-as-a-judge}: We use both GPT-4o and GPT-5 as judges to improve scoring credibility and enable cross-judge consistency. They score each summary on completeness, accuracy, coherence, conciseness, and overall quality, with three runs per sample averaged to reduce variance. 

\item \textbf{Atomic-claim precision/recall}: GPT-4o extracts atomic factual claims from reference and generated summaries, and computes precision, recall, and F1 via semantic matching. Compared with holistic judge scores, this metric offers finer-grained measurement of factuality (precision) and completeness (recall), and does not require access to full source documents at evaluation time~\cite{zhang2025comprehensivesurveyprocessorientedautomatic}.

\item \textbf{Image-Text Alignment}: We generate captions for document images and compute semantic similarity between summary segments and captions using BGE-M3~\cite{chen2024bge} (threshold 0.65), then report recall~\cite{hua2025v2xum}. ITA measures whether key visual evidence is faithfully reflected in the summary.

\item \textbf{ROUGE}: ROUGE-1, ROUGE-2, and ROUGE-L measure n-gram overlap with the reference summary, providing a lightweight indicator of coverage and surface text quality.
\end{itemize}

Image-Text Alignment and ROUGE are also used as reward components in the GRPO stage (Section~4.3). To ensure gains come from real quality improvement rather than reward fitting, we treat LLM-as-a-judge scores and atomic-claim precision/recall as independent validation metrics and exclude them from training objectives. When improvements in ITA and ROUGE are accompanied by stable gains in judge scores and atomic-claim metrics, this jointly verifies genuine multidimensional quality improvement rather than metric gaming.

\subsection{Experimental Setup}
\textbf{Baselines.}
We conduct comparative experiments on the MMLDSum-Bench benchmark, covering representative closed-source and open-source multimodal models, and build backbone-matched baselines to ensure fair comparison.  
The closed-source group includes GPT-5, Claude-4-Sonnet, Doubao-Seed-1.6, Qwen-VL-Max, Qwen3-VL-Plus, and Step-1o-Vision-32k.  
The open-source group includes strong community baselines across different scales and architectures: Qwen2.5-VL, Qwen3-VL, InternVL3.5, Gemma3, and Phi-4-Multimodal-Instruct, spanning lightweight to large-parameter settings for different deployment scenarios.
To avoid evaluation bias, Gemini-2.5-Pro and GPT-4o are excluded, since they are already used in our data construction and evaluation pipeline (Section~3 and Section~5.1).  
To verify the effectiveness of our two-stage training framework, we build SFT-only baselines on open-source backbones, including Qwen2.5-VL (3B/7B) and Qwen3-VL (8B), and compare them directly with MMLDSum-LLM.  
All models are evaluated under identical settings: the same test split, length-control strategy, prompt template, and a unified automated evaluation script for all metrics, ensuring fair and comparable results.

\textbf{Prompting and decoding.}
For all models, we use a unified instruction template that (i) asks for a concise global summary, (ii) explicitly requests grounding to both text and figures, and (iii) constrains output length.
For fair comparison, we enforce the same maximum output token budget and use deterministic decoding (temperature $=0$) unless a model requires sampling.

\textbf{Implementation details.}
For SFT, we train the model for 3 epochs using the AdamW optimizer with a learning rate of $5\times10^{-6}$ and a batch size of 1.
For GRPO, we employ a group size of $G=5$, a clipping parameter of $\epsilon=0.01$, and a KL regularization coefficient of $\eta=0.01$. The model is trained for 15 epochs with a learning rate of $10^{-6}$. All experiments are conducted on 8 NVIDIA H20 GPUs.

\begin{table*}[!t]
  \caption{Ablation study of MMLDSum-LLM. The final model is highlighted and annotated with improvement over the SFT baseline.}
  \label{tab:ablation}
  \centering
  \setlength{\tabcolsep}{1pt}
  \renewcommand{\arraystretch}{0.9}
  \small
  \resizebox{\textwidth}{!}{%
  \begin{tabular}{l ccccc ccccc ccc c ccc}
    \toprule
    \multicolumn{1}{c}{Variant} &
    \multicolumn{5}{c}{GPT-4o score} &
    \multicolumn{5}{c}{GPT-5 score} &
    \multicolumn{3}{c}{Atomic claim} &
    \multicolumn{1}{c}{ITA} &
    \multicolumn{3}{c}{ROUGE} \\
    \cmidrule(lr){2-6} \cmidrule(lr){7-11} \cmidrule(lr){12-14} \cmidrule(lr){15-15} \cmidrule(lr){16-18}
    & Comp. & Acc. & Conc. & Coh. & Overall & Comp. & Acc. & Conc. & Coh. & Overall & P & R & F1 & ITA-R & R-1 & R-2 & R-L \\
    \midrule
    qwen2.5-vl-7b-sft &
    \n{3.48} & \n{4.22} & \n{4.51} & \n{4.44} & \n{3.86} &
    \n{2.43} & \n{2.20} & \n{3.45} & \n{3.87} & \n{2.42} &
    \n{0.69} & \n{0.46} & \n{0.53} &
    \n{0.72} & \n{0.41} & \n{0.14} & \n{0.20} \\
    qwen2.5-vl-7b-sft (image\_weight, \textbf{I}) &
    \n{3.52} & \n{4.25} & \n{4.48} & \n{4.43} & \n{3.87} &
    \n{2.45} & \n{2.26} & \n{3.42} & \n{3.89} & \n{2.41} &
    \n{0.71} & \n{0.51} & \n{0.56} &
    \n{0.73} & \n{0.40} & \n{0.14} & \n{0.19} \\
    qwen2.5-vl-7b-sft (keywords\_weight,\textbf{K}) &
    \n{3.55} & \n{4.31} & \n{4.54} & \n{4.52} & \n{3.91} &
    \n{2.51} & \n{2.24} & \n{3.56} & \n{4.01} & \n{2.46} &
    \n{0.73} & \n{0.51} & \n{0.57} &
    \n{0.74} & \n{0.44} & \n{0.16} & \n{0.22} \\
    qwen2.5-vl-7b-sft (\textbf{I+K}) &
    \n{3.57} & \n{4.29} & \n{4.51} & \n{4.50} & \n{3.89} &
    \n{2.54} & \n{2.28} & \n{3.54} & \n{3.94} & \n{2.47} &
    \n{0.70} & \n{0.48} & \n{0.55} &
    \n{0.75} & \n{0.42} & \n{0.15} & \n{0.21} \\
    qwen2.5-vl-7b-sft + grpo &
    \n{3.78} & \n{4.61} & \n{4.58} & \n{4.74} & \n{4.07} &
    \n{2.63} & \n{2.45} & \n{3.70} & \n{4.10} & \n{2.61} &
    \n{0.70} & \n{0.54} & \n{0.59} &
    \n{0.83} & \n{0.49} & \n{0.19} & \n{0.24} \\
    MMLDSum-qwen2.5-vl-7b(ours) &
    \textbf{\n{3.82}} & \textbf{\n{4.65}} & \textbf{\n{4.60}} & \textbf{\n{4.79}} & \textbf{\n{4.13}} &
    \textbf{\n{2.63}} & \textbf{\n{2.47}} & \n{3.66} & \n{4.01} & \n{2.58} &
    \textbf{\n{0.71}} & \textbf{\n{0.54}} & \textbf{\n{0.59}} &
    \textbf{\n{0.87}} & \textbf{\n{0.51}} & \textbf{\n{0.21}} & \textbf{\n{0.26}} \\
    \midrule
       $\uparrow$(\%)  & \textbf{9.77} & \textbf{10.19} & \textbf{2.00} & \textbf{7.88} & \textbf{6.99} & \textbf{8.23} & \textbf{12.27} & \textbf{6.09} & \textbf{3.62} & \textbf{6.61} & \textbf{2.90} & \textbf{17.39} & \textbf{11.32} & \textbf{20.83} & \textbf{24.39} & \textbf{50.00} & \textbf{30.00} \\
    \bottomrule
  \end{tabular}
  }
\end{table*}

\subsection{Quantitative Results}
Table~\ref{tab:main} presents the comprehensive evaluation on MMLDSum-Bench across all four metric families: LLM-as-a-judge scoring from GPT-4o and GPT-5 (completeness, accuracy, conciseness, coherence, and overall), atomic-claim recall/F1, image-text alignment (ITA-R), and ROUGE.
Overall, MMLDSum-LLM consistently improves key-information coverage and cross-modal consistency, with the largest gains on dimensions that directly reflect \emph{completeness} (GPT-4o/GPT-5 completeness and atomic recall) and \emph{visual evidence alignment} (ITA-R).

\textbf{MMLDSum-qwen3vl-8b achieves open-source SOTA and approaches top closed-source models.}
MMLDSum-qwen3vl-8b reaches a GPT-4o overall score of $\n{4.51}$ and a GPT-5 overall score of $\n{3.21}$, surpassing all open-source baselines and approaching leading closed-source systems (Claude-4-Sonnet: $\n{4.52}$/$\n{3.58}$; Step-1o-Vision-32k: $\n{4.57}$/$\n{3.51}$).
On atomic-claim recall---the direct signal of factual completeness---our model achieves $\n{0.85}$, approaching GPT-5 ($\n{0.90}$) and substantially outperforming all other closed-source models (next best: Qwen3-VL-Plus and Doubao-Seed-1.6 at $\n{0.71}$).

\textbf{Two-stage training yields consistent gains across all backbone sizes.}
On Qwen3-VL-8B, the two-stage framework raises GPT-4o overall from $\n{4.29}$ to $\n{4.51}$ ($+5.1\%$), GPT-5 overall from $\n{2.78}$ to $\n{3.21}$ ($+15.5\%$), and atomic recall from $\n{0.73}$ to $\n{0.85}$ ($+16.4\%$).
On Qwen2.5-VL-7B, GPT-4o completeness improves by $+9.8\%$ ($\n{3.48}\!\rightarrow\!\n{3.82}$), GPT-4o overall by $+7.0\%$ ($\n{3.86}\!\rightarrow\!\n{4.13}$), and ITA-R by $+20.8\%$ ($\n{0.72}\!\rightarrow\!\n{0.87}$); the 7B model surpasses Qwen3-VL-32B on ITA-R ($\n{0.87}$ vs.\ $\n{0.78}$) with four times fewer parameters.
Even on the 3B backbone, GPT-4o overall gains $+20.3\%$ ($\n{3.15}\!\rightarrow\!\n{3.79}$) and ITA-R improves by $+52.0\%$ ($\n{0.50}\!\rightarrow\!\n{0.76}$), exceeding Qwen2.5-VL-32B on ITA-R ($\n{0.76}$ vs.\ $\n{0.63}$).

\textbf{Closed-source models lead on judge scores, yet coverage gaps persist across all systems.}
Closed-source models achieve consistently high judge scores (GPT-4o overall: $\n{4.52}$--$\n{4.83}$), but atomic-claim recall lags precision across most models---even GPT-5 ($\n{0.90}$ aggregate recall) degrades at 64k tokens (Appendix~D)---confirming that fully faithful long-context summarization remains an open problem.

SFT-only baselines still exhibit omissions and cross-modal inconsistencies, reflecting the limits of token-level cross-entropy on sparse evidence.
Both judges yield convergent rankings (Overall Spearman $\rho{=}0.894$, see Table~\ref{tab:judge-agreement} in Appendix~D;GPT-5 applies a stricter standard); boundary cases include \textit{Phi-4-Multimodal-Instruct} ($\n{1.76}$, limited Chinese capability) and \textit{Step-1o-Vision-32k} (32k context ceiling).
Length-stratified heatmaps (Appendix~D, Figures~\ref{fig:appendix-d-atomic}--\ref{fig:appendix-d-gpt5}) confirm that MMLDSum-LLM's advantage is most pronounced in the $16$k--$64$k range, where it achieves the best trade-off among open-source models across all four metric families.

\subsection{Ablation Study}
Table~\ref{tab:ablation} validates the contribution of each component on the Qwen2.5-VL-7B backbone.
Visual-alignment weighting (\textbf{I}) primarily boosts cross-modal consistency (ITA-R: $\n{0.72}\!\rightarrow\!\n{0.73}$, $+1.4\%$), while keyword-aware weighting (\textbf{K}) primarily improves key-fact retention (atomic recall: $\n{0.46}\!\rightarrow\!\n{0.51}$, $+10.9\%$). Effects are not isolated: \textbf{K} also lifts ITA-R to $\n{0.74}$, and \textbf{I} also benefits atomic precision. Combining both (\textbf{I+K}) further raises ITA-R to $\n{0.75}$.
Adding GRPO yields substantially larger sequence-level gains—ITA-R improves to $\n{0.83}$ ($+15.3\%$ over baseline)—by directly optimizing summary-level objectives that token-level cross-entropy cannot enforce. GRPO synergizes with weighted SFT rather than acting as a standalone boost.
Combining all components yields the best trade-off: ITA-R $\n{0.72}\!\rightarrow\!\n{0.87}$ ($+20.8\%$) and GPT-4o overall $\n{3.86}\!\rightarrow\!\n{4.13}$ ($+7.0\%$), exceeding any individual component (Table~\ref{tab:ablation}).


\section{Discussion}

\textbf{Token-level weighting and sequence-level rewards jointly target the two core failure modes.} In Stage~1, keyword-aware and visual-alignment weighting counter key-information omission and cross-modal hallucination by raising the gradient on salient entities and visually grounded spans. Stage~2 reinforces the same two axes at the sequence level: the keyword-coverage reward penalizes missing entities, and the image-text alignment reward suppresses ungrounded visual mentions, especially on image-dominant documents.

\textbf{Composite reward balances coverage and faithfulness without sacrificing conciseness or coherence.} Optimizing a single reward in isolation over-shoots one axis at the cost of others: keyword coverage alone inflates length with peripheral entities, and image-text alignment alone encourages indiscriminate visual mentions. Coupling these signals with ROUGE and a length penalty lets the four components mutually regularize, yielding summaries that are informative, visually faithful, concise, and coherent.



\section{Conclusion}
We study multimodal long-document summarization under long-context and cross-modal evidence sparsity, focusing on key-information omission and cross-modal hallucination.
We introduce MMLDSum-Bench, a multi-domain, multi-length, multi-ratio benchmark, and propose MMLDSum-LLM, a two-stage recipe that combines weighted SFT (visual alignment and keyword awareness) with GRPO using verifiable, multi-objective rewards.
Across automatic and judge-based evaluations, MMLDSum-LLM improves key-information coverage and cross-modal consistency compared with SFT-only baselines.
Future work includes stronger chart-specific grounding, adaptive reward re-weighting conditioned on length/ratio, and more reliable multimodal evaluation protocols.

\clearpage
\section*{Limitations}
First, our rewards still rely on proxy signals (e.g., ROUGE, keyword coverage, and image-text alignment) that can miss fine-grained factual errors or chart-specific reasoning, especially for dense plots and complex diagrams. Second, long-context behavior remains fragile: when key evidence is sparse and distributed across distant sections, the model may still omit crucial details or overfit local evidence despite weighted training. Third, evaluation costs remain high because judge-based scoring and atomic-claim verification are computationally expensive, which limits large-scale ablations and rapid iteration. Finally, our benchmark focuses on static documents with pre-extracted images; extending to dynamic or interactive visuals (e.g., videos or embedded charts with underlying data) remains future work.

\section*{Reproducibility}
\label{sec:repro}
To support full reproducibility and community adoption, we will publicly release: (i)~\textbf{MMLDSum-Bench}, including all benchmark documents, paired reference summaries, and split metadata (SFT/RL/test); (ii)~\textbf{training code} for both Stage~1 (visual-alignment and keyword-aware weighted SFT) and Stage~2 (GRPO with multi-objective verifiable rewards), together with training configuration files and hyperparameter settings used in all reported experiments; (iii)~\textbf{evaluation code}, covering the full automated evaluation suite---LLM-as-a-judge prompts (GPT-4o and GPT-5 five-dimension scoring), atomic-claim extraction and verification pipelines, ITA-R computation (BGE-M3 with threshold 0.65), and ROUGE scoring; (iv)~\textbf{model checkpoints} for all reported MMLDSum-LLM variants (3B, 7B, 8B); and (v)~all \textbf{inference prompt templates} used during model evaluation. All training runs use fixed random seeds. We will document software versions (Python, PyTorch, Transformers, vLLM) and hardware specifications (NVIDIA H20 $\times$ 8). Benchmark data is filtered to remove personally identifiable information, and all source dataset licenses are respected.

\section*{Use of AI Assistants}
The AI assistant, GPT-4o, is used solely for refining the writing of our paper.

\bibliography{mmldsum_emnlp2026}

\iffullappendix
\appendix

\section{Dataset Construction and Statistics}
\subsection{MMLDSum-Bench Statistics (Illustrative Figure)}
\begin{figure}[!tbp]
  \centering
  \begin{subfigure}{\columnwidth}
    \centering
    \includegraphics[width=\linewidth]{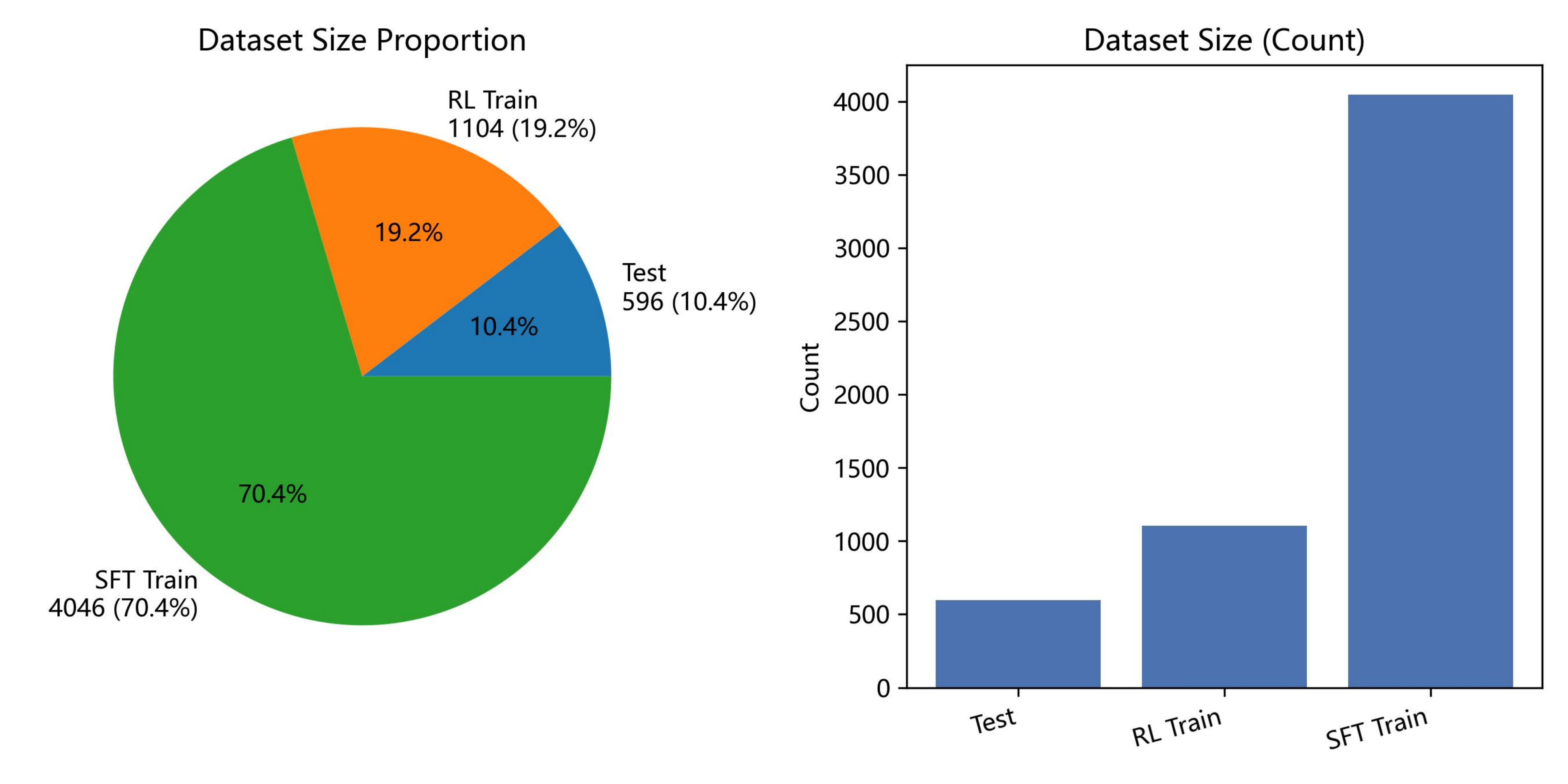}
    \caption{Dataset size and proportion.}
  \end{subfigure}

  \medskip
  \begin{subfigure}{\columnwidth}
    \centering
    \includegraphics[width=\linewidth]{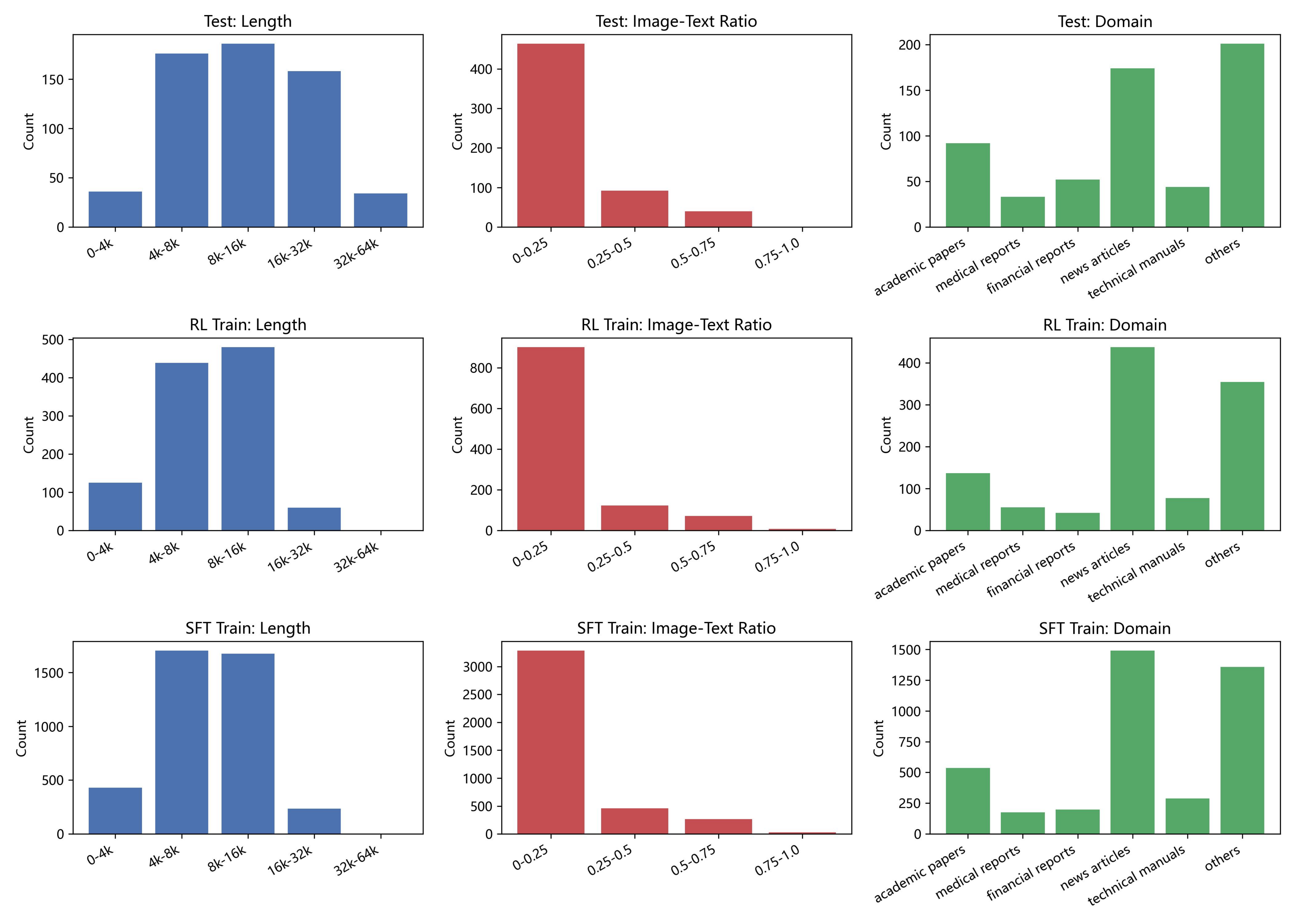}
    \caption{Length, ratio, and domain distributions.}
  \end{subfigure}
  \caption{MMLDSum-Bench statistics for the SFT, RL, and test splits.}
  \label{fig:appendix-b1}
\end{figure}

Figure~\ref{fig:appendix-b1} summarizes the data composition and distribution patterns of MMLDSum-Bench. The split size overview highlights the relative scale of the SFT, RL, and test sets. The distribution grid reports (i) length bins following the paper's standard 4k/8k/16k/32k/64k ranges, (ii) image--text ratio buckets at 0.25/0.5/0.75/1.0, and (iii) the six-domain taxonomy used in the paper: academic papers, medical reports, financial reports, news articles, technical manuals, and others. Together, these statistics validate that the benchmark spans diverse domains and modality balances while remaining focused on long-context settings.

\subsection{Pipeline for Automatic Summary Construction}
\begin{figure*}[!tbp]
  \centering
  \includegraphics[width=\textwidth]{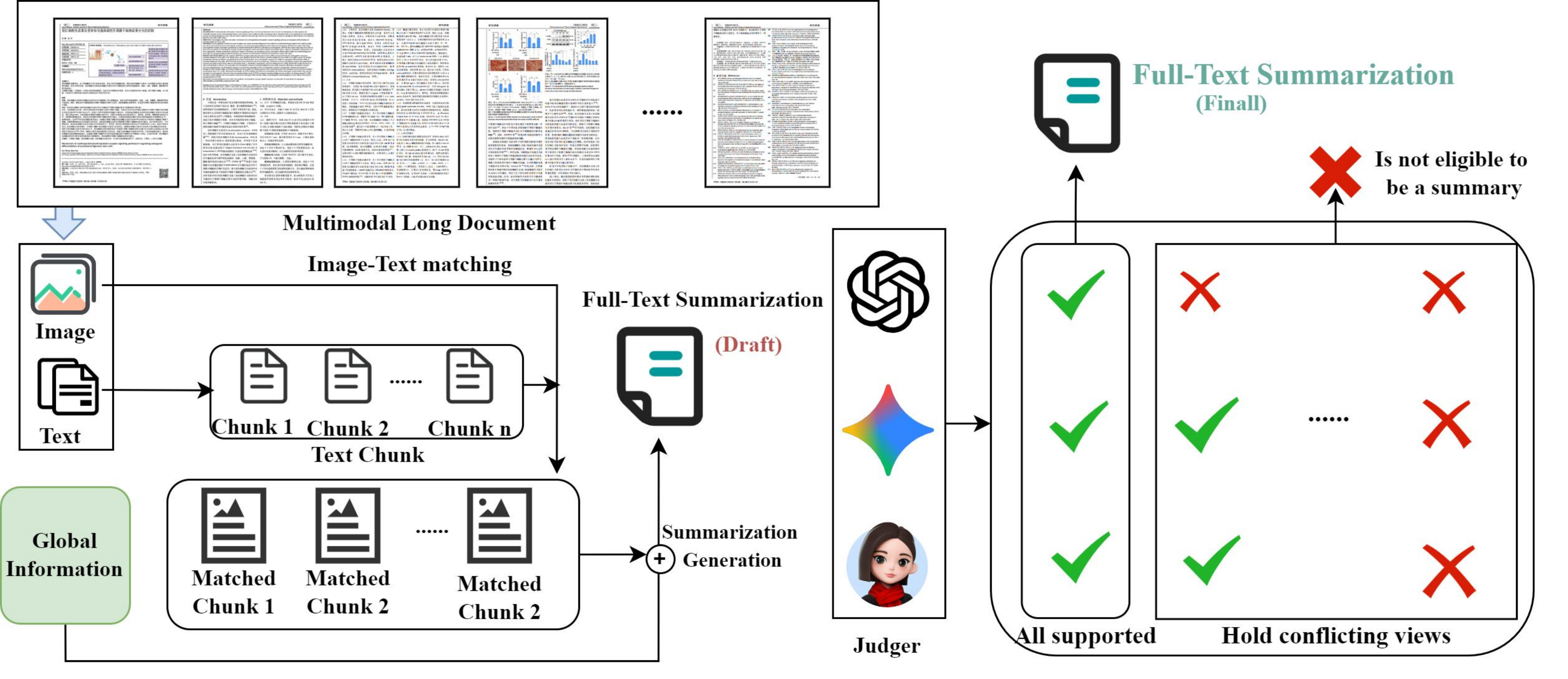}
  \caption{Pipeline for MMLDSum-Bench construction: image--text matching and chunking, global information fusion to draft summaries, and multi-model judging to select the final summary.}
  \label{fig:framework_data}
\end{figure*}

\subsection{Data Quality Validation Protocol and Results}
\label{sec:appendix-quality-validation}

\subsubsection{Annotator Profiles}
Three annotators (A1, A2, A3) participate in the quality validation. All hold graduate-level degrees in natural language processing or related fields and have at least two years of research experience with text summarization and multimodal document understanding. Prior to formal annotation, all annotators complete a calibration session on 30 pilot samples (excluded from the final evaluation set) to align scoring criteria and resolve ambiguities in dimension definitions.

\subsubsection{Stratified Sampling Protocol}
We draw 600 summaries from the benchmark via stratified sampling along three axes to ensure representative coverage:
\begin{itemize}[leftmargin=*]
\item \textbf{Domain}: samples are allocated proportionally across six domains (academic papers, medical reports, financial reports, news articles, technical manuals, and others).
\item \textbf{Context-length scale}: samples are drawn from all five length bins (4K, 8K, 16K, 32K, 64K tokens) with proportional allocation reflecting the benchmark distribution, while enforcing a minimum of 30 samples per bin.
\item \textbf{Visual-textual modality distribution}: samples cover all four modality categories (heavily text-dominant, lightly text-dominant, lightly image-dominant, heavily image-dominant), with oversampling applied to minority categories to ensure at least 20 samples per category.
\end{itemize}

\subsubsection{Annotation Scheme}
Each of the 600 summaries is independently scored by all three annotators on five dimensions using a 1--5 Likert scale:
\begin{itemize}[leftmargin=*]
\item \textbf{Completeness}: whether the summary covers all salient information from the source document across modalities.
\item \textbf{Accuracy}: whether the factual claims in the summary are correct and free of hallucinations.
\item \textbf{Coherence}: whether the summary is logically organized and easy to follow.
\item \textbf{Conciseness}: whether the summary avoids redundancy and unnecessary detail.
\item \textbf{Overall quality}: a holistic assessment of the summary.
\end{itemize}
In addition, 200 atomic claims are randomly sampled from the generated summaries. Each claim is independently verified by all three annotators against the source document (text and associated images) and labeled as \emph{Supported}, \emph{Partially Supported}, or \emph{Unsupported}.

\subsubsection{Detailed Annotation Results}

\paragraph{Per-dimension summary-level evaluation.}
Table~\ref{tab:annotation-results} reports per-dimension mean scores with 95\% bootstrap confidence intervals and pairwise Cohen's Kappa averaged over the three annotator pairs.

\begin{table}[t]
\centering
\small
\caption{Per-dimension human evaluation results on 600 stratified summaries.}
\label{tab:annotation-results}
\begin{tabular}{lccc}
\toprule
Dimension & Mean & 95\% CI & Cohen's $\kappa$ \\
\midrule
Completeness & 4.58 & [4.52, 4.64] & 0.80 \\
Accuracy      & 4.82 & [4.78, 4.86] & 0.86 \\
Coherence     & 4.75 & [4.70, 4.80] & 0.84 \\
Conciseness   & 4.68 & [4.62, 4.74] & 0.82 \\
Overall       & 4.70 & [4.65, 4.75] & 0.83 \\
\bottomrule
\end{tabular}
\end{table}

\paragraph{Claim-level manual verification.}
Table~\ref{tab:claim-verification} reports the results of claim-level verification on 200 atomic claims, along with inter-annotator agreement measured by Fleiss' Kappa.

\begin{table}[t]
\centering
\small
\caption{Claim-level manual verification results on 200 atomic claims.}
\label{tab:claim-verification}
\begin{tabular}{lc}
\toprule
Metric & Value \\
\midrule
Supported rate        & 88.5\% \\
Partially supported rate & 7.0\% \\
Unsupported rate      & 4.5\% \\
Annotator agreement (Fleiss' $\kappa$) & 0.81 \\
\bottomrule
\end{tabular}
\end{table}

\paragraph{Regeneration-effect analysis.}
Table~\ref{tab:regen-effect} quantifies the effect of the regeneration module by comparing quality scores before and after regeneration for the subset of samples that triggered the quality-threshold filter.

\begin{table}[t]
\centering
\small
\caption{Effect of the regeneration module on filtered samples.}
\label{tab:regen-effect}
\begin{tabular}{lccc}
\toprule
Dimension & Before & After & $\Delta$ \\
\midrule
Completeness & 3.42 & 4.51 & +1.09 \\
Accuracy      & 3.78 & 4.76 & +0.98 \\
Coherence     & 3.85 & 4.70 & +0.85 \\
Conciseness   & 3.90 & 4.62 & +0.72 \\
Overall       & 3.56 & 4.65 & +1.09 \\
\midrule
Pass rate     & \multicolumn{3}{c}{37.2\% $\rightarrow$ 91.8\%} \\
\bottomrule
\end{tabular}
\end{table}

The results confirm that the dataset maintains high annotation quality across all dimensions (overall mean: 4.7/5.0; overall Cohen's $\kappa$: 0.83). Claim-level verification indicates a low unsupported-claim rate (4.5\%), and the regeneration mechanism yields substantial quality improvements (average score increase of +0.95 across dimensions; pass-rate improvement from 37.2\% to 91.8\%).

\subsection{Prompts for Automatic Summary Construction}
The following prompt templates are used in the three-stage automatic summary construction pipeline described in Section~3.

\begin{tcolorbox}[promptboxstyle,title=Chunk-Level Summary Generation Prompt (System)]
\begin{Verbatim}
Role:
You are an expert in full-information multimodal summarization. Generate summaries in Simplified Chinese. Your core objective is to preserve all source information, align correctly with image positions, jointly present text and image content, and keep the summary logic/order exactly consistent with the source.

Task background and objective:
- Input data: The user provides source text and an image list. In the source text, <image X> (X is a number) denotes an image marker. Marker order matches the image list order one-to-one (e.g., <image 1> corresponds to image 1 in the list).
- Summary requirement: You must summarize both source text and all images. Do not omit any textual details (background, causes, process, conclusions, opinions, definitions, features, data, time, cases, etc.) or image information. Do not repeat content.
- Image recall requirement: The summary must recall all source images. Keep all markers like <image 1> at their corresponding source positions. Do not modify or delete these markers. Ensure every marker in the summary has a matching image in the source.

Summary rules:
1. Completeness:
- Reproduce details sentence by sentence so users can recover all source information without loss.
- Fully extract image information by image type (chart/diagram/scene/flowchart/text-in-image), including key elements, data, relations, and scene descriptions; integrate naturally at corresponding positions.
2. Accuracy:
- All content (events, opinions, data, time, wording) must come from the source. No fabrication.
- Keep critical wording exactly consistent with the source (e.g., if the source says "less than 3%", do not rewrite it as "only 3%" or "more than 3%").
- Preserve relative time expressions (e.g., "this year", "last month", "the first half of the year"); do not convert them into absolute dates.
- Keep summary logic and order exactly consistent with the source.
3. Image content presentation:
- Text and image information should have equal importance, both presented completely in source order with natural transitions.

Notes:
1. Do not delete or modify any <image X> markers.
2. Do not reorder source content or logic.
3. Do not fabricate non-source content (text details, image info, data, or opinions).
4. Do not output non-summary notes (e.g., "Image details are integrated above.").
5. Do not simplify key source details.
\end{Verbatim}
\end{tcolorbox}

\begin{tcolorbox}[promptboxstyle,title=Global Information Extraction Prompt (System)]
\begin{Verbatim}
Role:
You are an information extraction specialist. Extract document-level global information from the given document and output it in the required JSON format to support downstream summarization.

JSON fields (must include all):
- "topic": one-sentence summary of the document's core topic
- "outline": a list of major section/paragraph titles
- "key_entities": repeatedly appearing key entities, including but not limited to people, organizations, locations, products, technologies, and concepts

Example output (strict JSON, no extra characters):
{
  "topic": "Global AI chip market analysis for Q3 2024",
  "outline": ["Market overview", "Major vendor updates", "Technology trends", "Outlook"],
  "key_entities": ["NVIDIA", "AMD", "H100", "compute power"]
}
\end{Verbatim}
\end{tcolorbox}

\begin{figure*}[!t]
  \centering
  \captionsetup[subfigure]{font=footnotesize,skip=3pt}
  \begin{subfigure}[b]{0.32\textwidth}
    \centering
    \includegraphics[width=\linewidth]{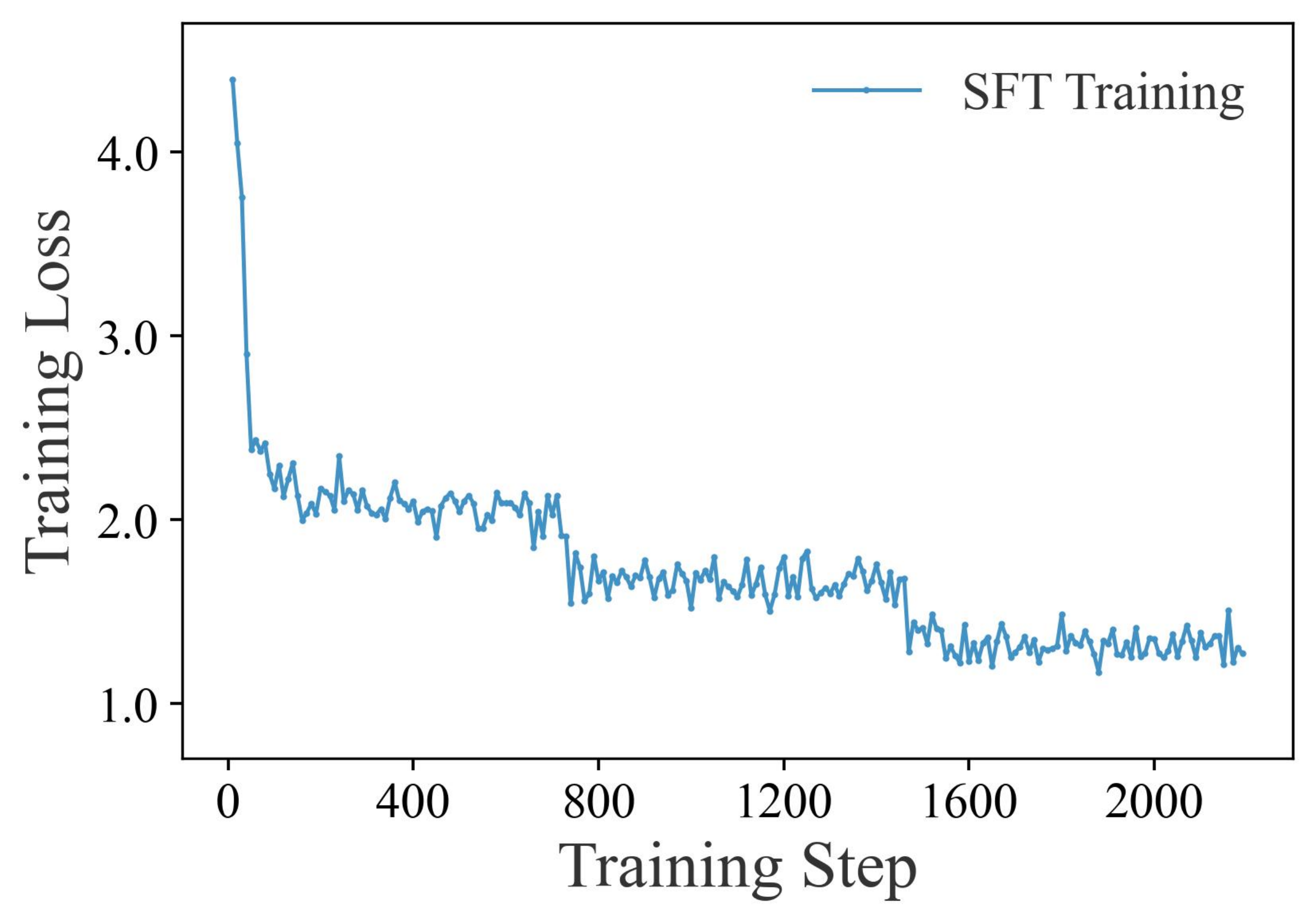}
    \caption{SFT loss.}
  \end{subfigure}\hfill
  \begin{subfigure}[b]{0.32\textwidth}
    \centering
    \includegraphics[width=\linewidth]{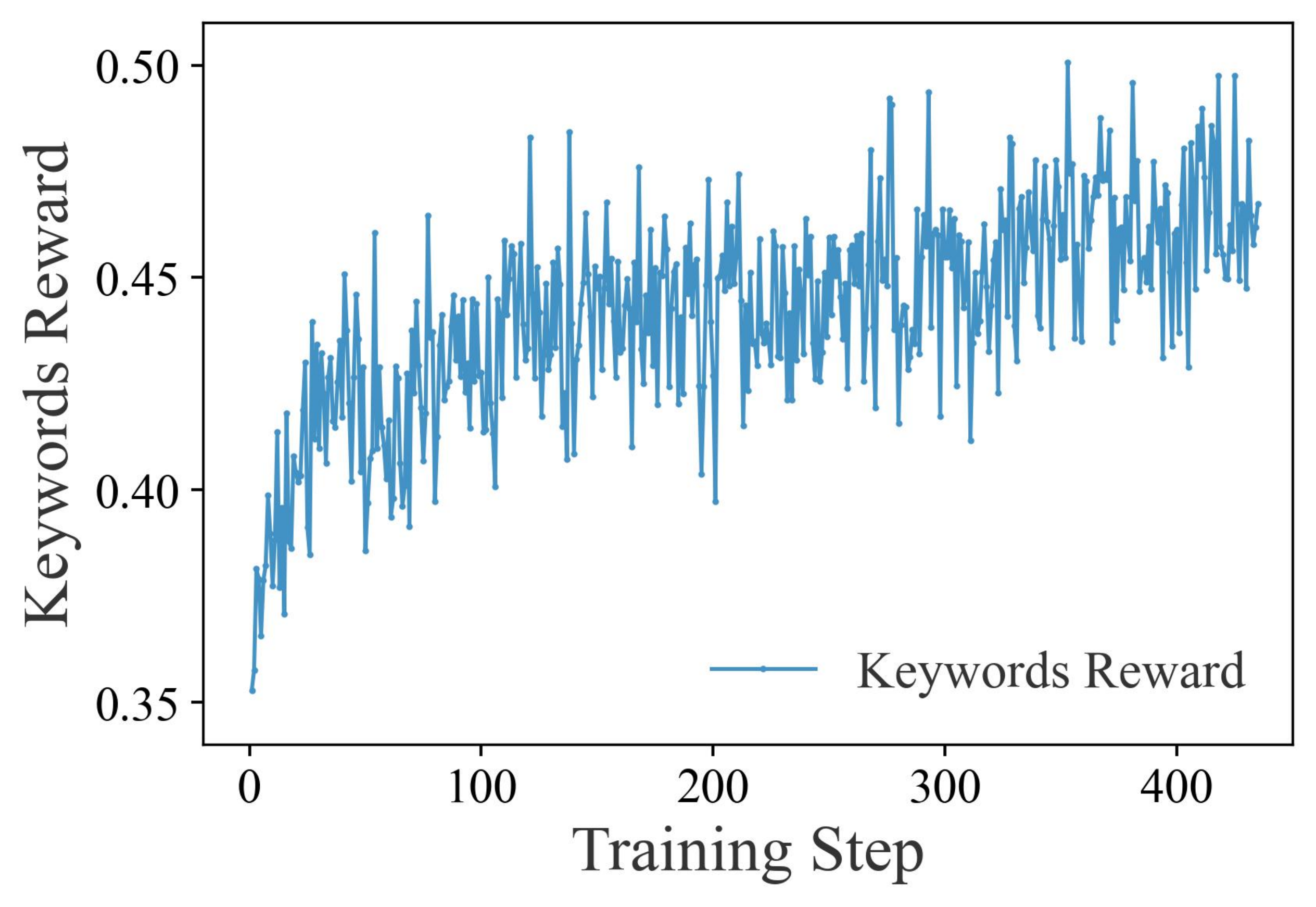}
    \caption{Keyword reward.}
  \end{subfigure}\hfill
  \begin{subfigure}[b]{0.32\textwidth}
    \centering
    \includegraphics[width=\linewidth]{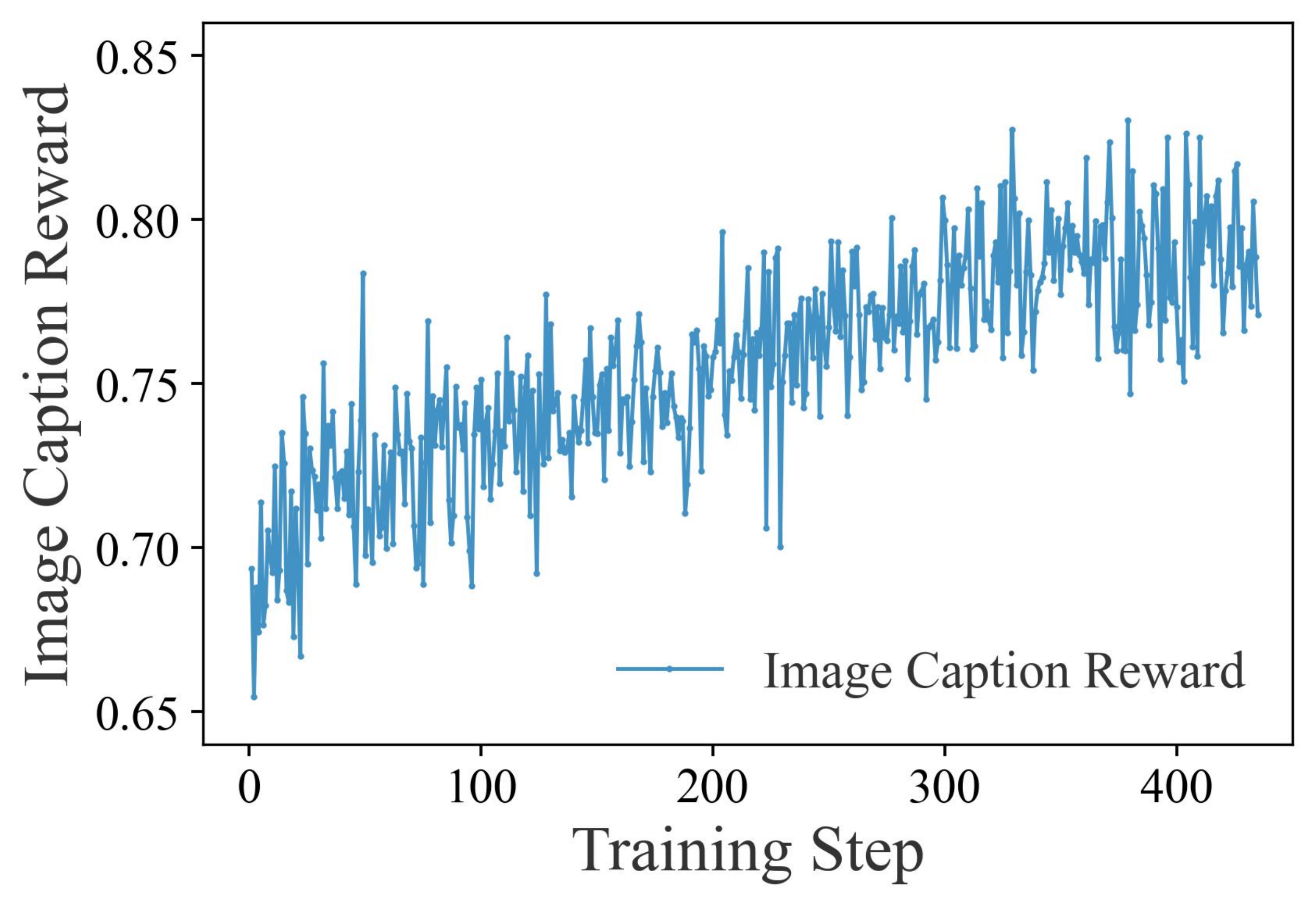}
    \caption{ITA reward.}
  \end{subfigure}

  \smallskip
  \begin{subfigure}[b]{0.32\textwidth}
    \centering
    \includegraphics[width=\linewidth]{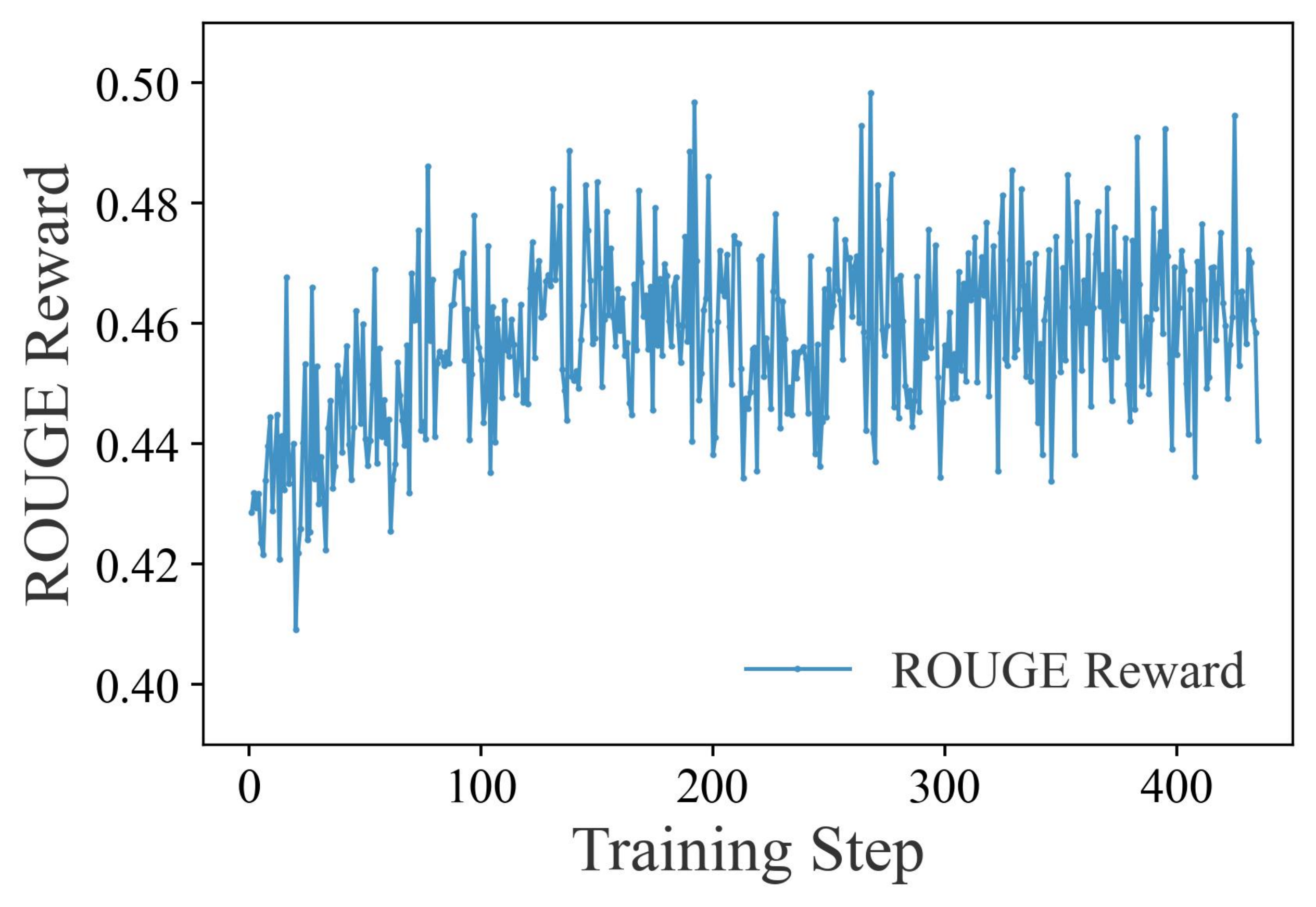}
    \caption{ROUGE reward.}
  \end{subfigure}\hfill
  \begin{subfigure}[b]{0.32\textwidth}
    \centering
    \includegraphics[width=\linewidth]{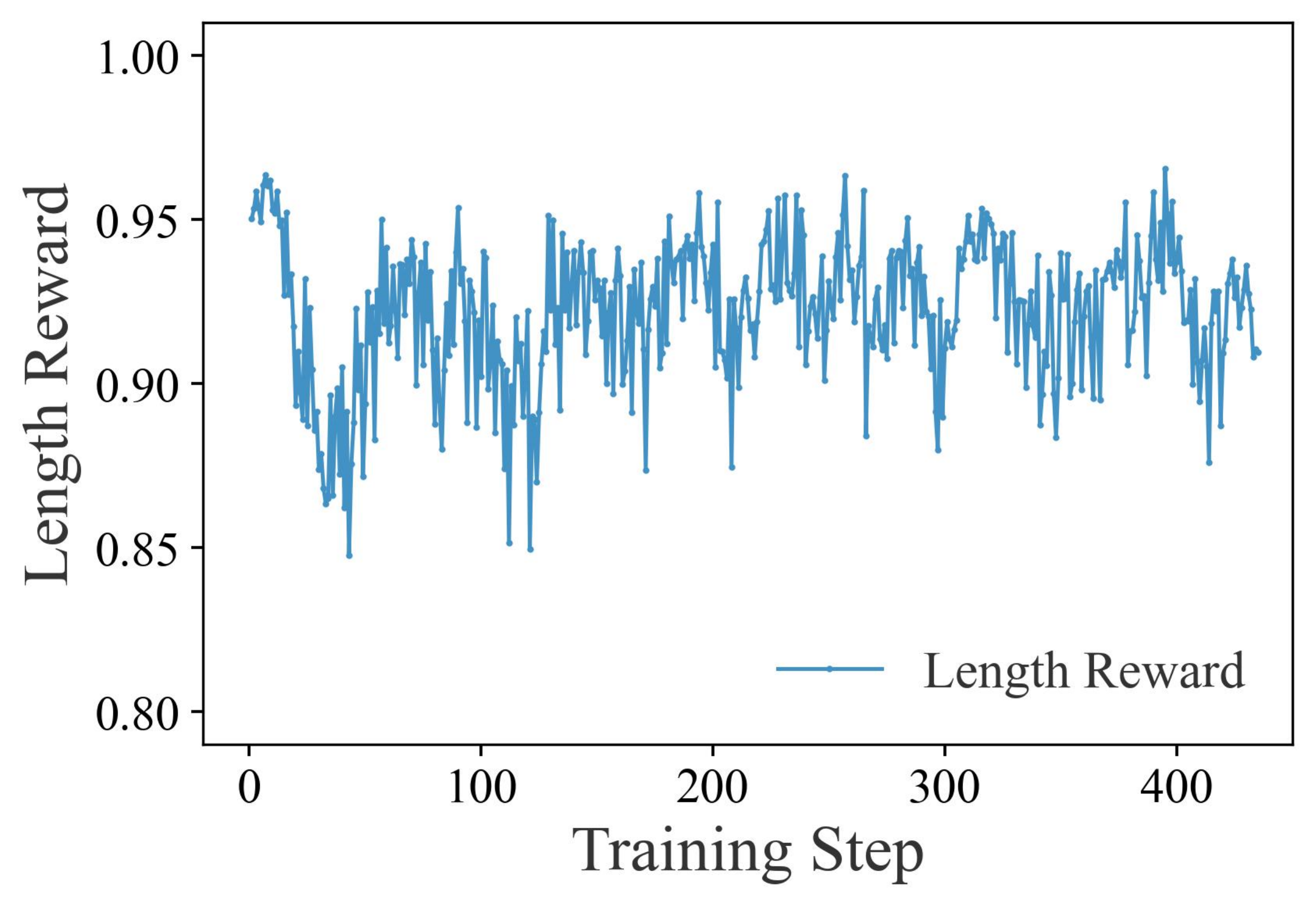}
    \caption{Length reward.}
  \end{subfigure}\hfill
  \begin{subfigure}[b]{0.32\textwidth}
    \centering
    \includegraphics[width=\linewidth]{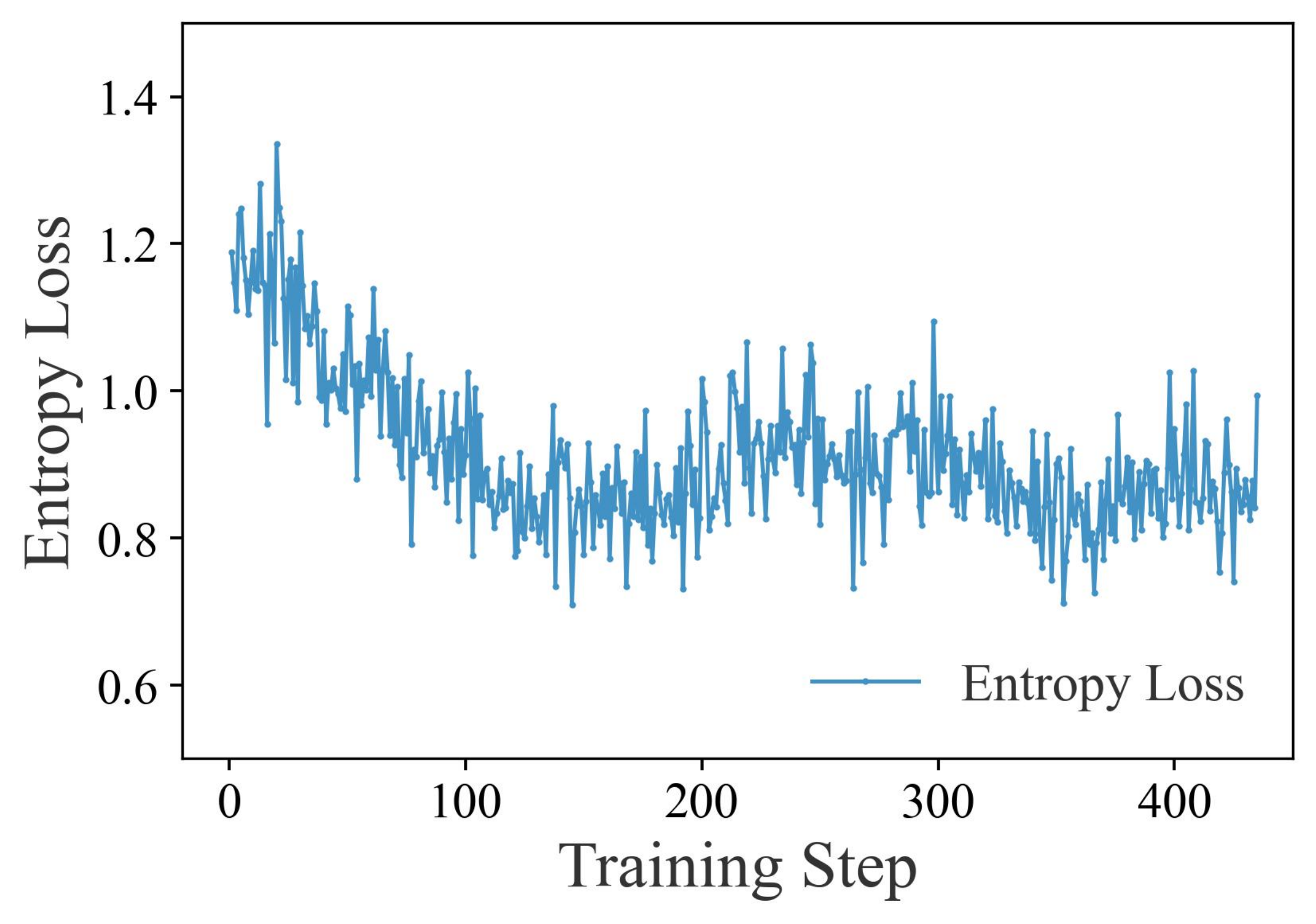}
    \caption{Entropy (exploration).}
  \end{subfigure}
  \caption{Training curves for SFT loss and GRPO reward signals.}
  \label{fig:training-curves}
\end{figure*}

\begin{tcolorbox}[promptboxstyle,title=Global Summary Generation Prompt (System)]
\begin{Verbatim}
Role:
You are a full-information replication summarization expert. You can process both text and image content jointly and generate summaries in Simplified Chinese. The summary must be complete and accurate, with logic and order exactly consistent with the source.

Task background and requirements:
- Input data: The user provides global document information and multiple chunk summaries.
  a) Global information includes topic, outline, and key entities, which helps reconstruct the source structure and avoid fragmented writing.
  b) In each chunk summary, <image X> marks images, and all markers have been globally reindexed in document order.
- Summary requirement: Summarize all chunk summaries and global information together, preserving content and order exactly, with no omission and no repetition.

Summary rules:
1. Structure: Use introduction - detailed bullet points - optional conclusion.
2. Content:
   a) Completeness: Reproduce all details from chunk summaries (background, causes, process, conclusions, definitions, features, data, time, cases, etc.).
   b) Accuracy:
   - No fabricated content.
   - Keep key wording consistent with source values and semantics.
   - Preserve relative time expressions.
   - Keep logic/order exactly consistent with the source.
   c) Image content:
   - Summarize both text and image content in source order.
   - Extract key image elements/data/relations and integrate naturally.
3. Format: Any information coming from images must be wrapped with <image_tag>...</image_tag>.

Workflow:
Step 1: Write one introductory paragraph summarizing text and images.
Step 2: Expand in ordered bullet points according to source sequence, using global topic/outline/entities to improve coherence between chunks. All image-derived content must be wrapped by <image_tag>...</image_tag>.
Step 3: Self-check coverage, factual consistency (especially data/time), and order consistency.
Step 4: If needed, add a final concluding paragraph.

Important constraints:
1. Do not reorder chunk summaries.
2. Keep text and image information balanced.
3. Ensure all content is source-grounded and consistent.
4. Do not output explanatory meta text.
5. Do not output raw image markers such as <image x> in the final summary.
\end{Verbatim}
\end{tcolorbox}

\section{Training Curves}

Figure~\ref{fig:training-curves} visualizes the optimization dynamics across supervised fine-tuning and GRPO. The SFT loss decreases steadily, while reward components (keyword, caption, ROUGE, and length) rise as the policy improves. The entropy curve indicates exploration during RL, which stabilizes as rewards converge.

\section{Prompts for Inference and Evaluation}
This appendix presents the prompt templates used for model inference and automated evaluation.

\begin{tcolorbox}[promptboxstyle,title=Inference-Time Summary Generation Prompt (User)]
\begin{Verbatim}
You are an expert in multimodal long-document summarization. Your task is to generate a summary in Chinese for a multimodal long document. The summary must be complete, accurate, and follow the same logical order as the source.

Task background:
The user provides source text and an image list. In the source text, <image X> (X is a number) is an image marker. Marker order maps one-to-one to the image list (e.g., <image 1> corresponds to image 1).
You must summarize both text and image content, and keep their presentation order exactly aligned with the source.

Summary requirements:
Use an introduction - detailed bullet points - optional conclusion structure:
1. Opening: one paragraph that gives a high-level overview of text and images;
2. Middle bullet points: expand in detail according to source order and paragraph hierarchy, accurately presenting each part's key content;
3. Ending (optional): one paragraph summarizing the main idea, core conclusions, and overall information.

=== Source Document Start ===
{article}
=== Source Document End ===

Now generate the summary based on the document and images. Output only the summary, and do not output any irrelevant content.
\end{Verbatim}
\end{tcolorbox}

\begin{tcolorbox}[promptboxstyle,title=Five-Dimension LLM-as-a-Judge Prompt (System)]
\begin{Verbatim}
Role:
You are a precise and professional image-text summary evaluator specialized in scoring Chinese summaries generated from text+image inputs. You provide rigorous step-by-step analysis and quantitative scores.

Task and output:
- Input includes source text, image list, and generated summary. <image X> markers in source map one-to-one to the image list.
- Score each dimension from 1 to 5: completeness, accuracy, conciseness, coherence, and overall quality.
- Output must include two parts:
  1) detailed reasoning process for each dimension;
  2) final JSON scores for automatic extraction.

Scoring dimensions:
1. Completeness: no missing core text info or key image info.
2. Accuracy: no factual deviation, contradiction, or fabrication in text/image descriptions.
3. Conciseness: no irrelevant content, redundancy, or repeated statements.
4. Coherence: clear ordering and logical flow consistent with source text-image structure.
5. Overall quality: holistic quality considering all dimensions.

Output format:
[Reasoning]
... detailed analysis for each dimension ...
[Scores] (JSON only for scores)
{"completeness": [score], "accuracy": [score], "conciseness": [score], "coherence": [score], "overall": [score]}

Important:
- In the final JSON, output numeric values only (e.g., 1, 2, 3, 4, 5), without units or extra text.
\end{Verbatim}
\end{tcolorbox}

\begin{tcolorbox}[promptboxstyle,title=Five-Dimension LLM-as-a-Judge Prompt (User)]
\begin{Verbatim}
=== Source Document Start ===
{article}
=== Source Document End ===

=== Summary Start ===
{summary}
=== Summary End ===
\end{Verbatim}
\end{tcolorbox}

\begin{tcolorbox}[promptboxstyle,title=Atomic-Claim Extraction Prompt (System)]
\begin{Verbatim}
Your task is to extract all independent atomic factual claims from the provided Chinese summary text. An atomic claim is the smallest complete statement that can be judged true or false.

Strict rules (must be followed 100%):
1. One sentence, one fact: each claim must contain exactly one independent fact.
2. Explicit information only: do not add inference, external knowledge, assumptions, interpretation, or opinion.
3. Preserve details: keep all dates, numbers, amounts, named entities, acronyms, and specific descriptions unchanged.
4. Split compound statements connected by words such as "and/or/also/includes" into multiple independent claims.
5. Split modifier-bearing facts into independent atomic claims when modifiers carry standalone facts.
6. Format each claim as a complete declarative sentence with proper punctuation.
7. No omission and no duplication.
8. Output must be a single valid JSON string only, with no prefix/suffix text.
   - Key name must be exactly: atomic_claims
   - No extra keys
   - Array elements must be JSON strings
   - Use ASCII JSON punctuation only

Required output format:
{"atomic_claims": ["Atomic claim 1.", "Atomic claim 2.", "Atomic claim 3."]}
\end{Verbatim}
\end{tcolorbox}

\begin{tcolorbox}[promptboxstyle,title=Atomic-Claim Extraction Prompt (User)]
\begin{Verbatim}
=== Summary Start ===
{summary}
=== Summary End ===
\end{Verbatim}
\end{tcolorbox}

\begin{tcolorbox}[promptboxstyle,title=Atomic-Claim Verification Prompt (System)]
\begin{Verbatim}
You are a factual verification expert. Determine whether each atomic factual claim is supported by the summary.

Decision rule:
- true: the summary explicitly contains or directly supports the claim
- false: the summary does not mention the claim or contradicts it

Output format (JSON only, no extra text):
{"results": {"1": true, "2": false, "3": true}}

Notes:
1. Output only claim IDs and boolean judgments; do not output claim text.
2. IDs must align one-to-one with the input claim numbering.
3. You must return judgments for all input claims.
\end{Verbatim}
\end{tcolorbox}

\begin{tcolorbox}[promptboxstyle,title=Atomic-Claim Verification Prompt (User)]
\begin{Verbatim}
Summary:
{summary}

Atomic claims:
{claims}

Output the support judgment for each atomic-claim ID.
\end{Verbatim}
\end{tcolorbox}

\section{Additional Statistics}

This section presents a detailed analysis of model performance stratified by context-length bin across five metric families.
The heatmaps in Figures~\ref{fig:appendix-d-gpt4o}--\ref{fig:appendix-d-gpt5} visualize per-model, per-length-bin performance, complementing the aggregate scores in Table~\ref{tab:main} and providing finer-grained insight into how summarization quality degrades (or is maintained) under increasing document length.
Length bins follow the standard 4k/8k/16k/32k/64k token ranges, and each cell reports the average score for all test documents in that bin.

\begin{figure}[!htbp]
  \centering
  \includegraphics[width=\columnwidth]{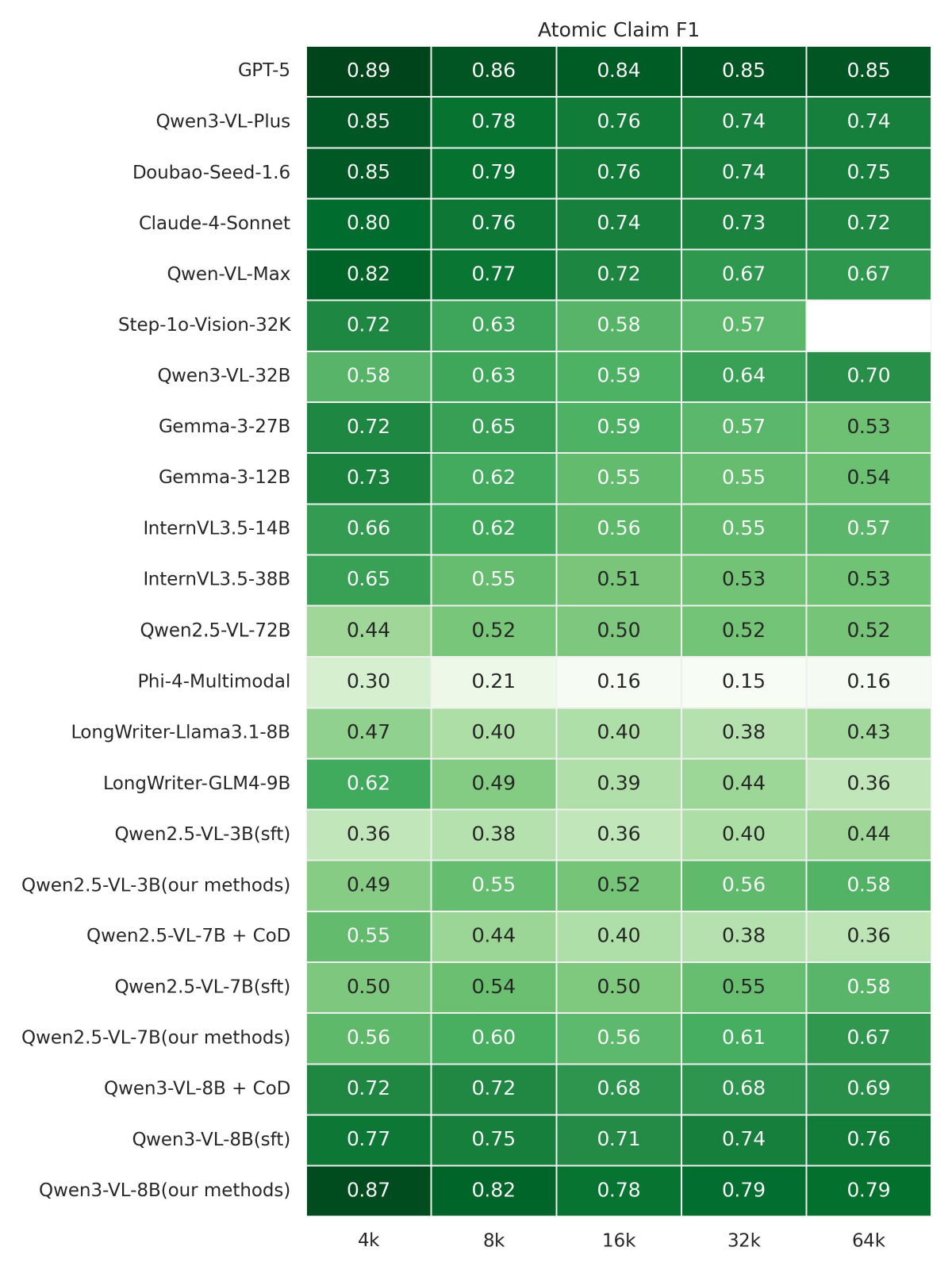}
  \caption{Atomic-claim F1 on MMLDSum-Bench. Higher values indicate better performance.}
  \label{fig:appendix-d-atomic}
\end{figure}

\paragraph{Atomic-claim F1 (Figure~\ref{fig:appendix-d-atomic}).}
Atomic-claim F1 is sensitive to document length: most models exhibit a clear downward trend as context length grows from 4k to 64k tokens, although the decrease is not strictly monotonic for every system.
The drop is most pronounced for weaker open-source baselines without explicit key-information training (e.g., Phi-4-Multimodal: $\n{0.30}\!\rightarrow\!\n{0.16}$; LongWriter-GLM4-9B: $\n{0.62}\!\rightarrow\!\n{0.36}$), and models with a 32k context ceiling (Step-1o-Vision-32K, InternVL3.5-14B/38B) likewise degrade visibly beyond 16k tokens; Step-1o-Vision-32K does not produce a result in the 64k bin due to forced truncation.
MMLDSum-qwen3vl-8b is the strongest open-source system in every length bin ($\n{0.87}, \n{0.82}, \n{0.78}, \n{0.79}, \n{0.79}$), surpassing both its backbone-matched SFT baseline (Qwen3-VL-8B-sft: $\n{0.77}, \n{0.75}, \n{0.71}, \n{0.74}, \n{0.76}$) and the CoD variant (Qwen3-VL-8B + CoD: $\n{0.72}, \n{0.72}, \n{0.68}, \n{0.68}, \n{0.69}$) across all bins, and remains stable around $\n{0.78}$--$\n{0.79}$ in the 16k--64k range, suggesting that keyword-aware weighted SFT together with GRPO's sequence-level keyword-coverage reward helps mitigate the difficulty of evidence selection in longer documents.
A small number of systems instead exhibit non-monotonic or mildly increasing F1 with length (e.g., Qwen3-VL-32B: $\n{0.58}$ at 4k vs.\ $\n{0.70}$ at 64k; Qwen2.5-VL-72B: $\n{0.44}\!\rightarrow\!\n{0.52}$), indicating that document length alone is not the sole determinant of atomic-claim quality and that each model's specific long-context behavior also plays a role.

\begin{figure}[!htbp]
  \centering
  \includegraphics[width=0.9\columnwidth]{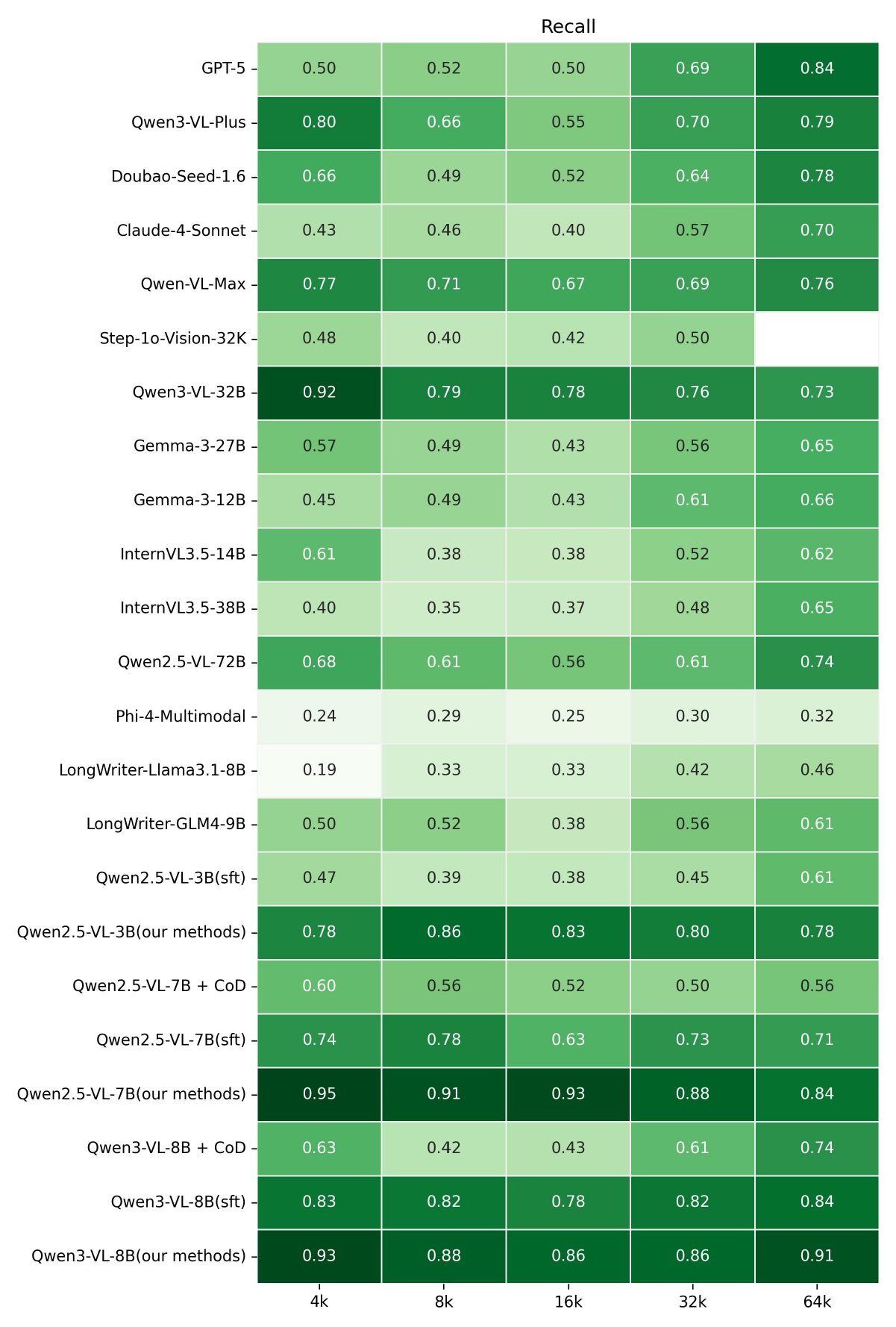}
  \caption{ITA-R on MMLDSum-Bench. Higher values indicate better performance.}
  \label{fig:appendix-d-ita}
\end{figure}

\paragraph{ITA-R (Figure~\ref{fig:appendix-d-ita}).}
ITA-R exhibits the strongest sensitivity to the visual-textual modality distribution and document length of any metric in our suite.
For most models, ITA-R is notably higher in the 4k--8k bin (where visual evidence is densely concentrated and spatially close to its textual descriptions) than in the 32k--64k bin (where images are scattered across distant document sections).
This degradation is particularly sharp for models without explicit visual-alignment training, confirming the theoretical motivation of our visual-alignment weighted loss.
MMLDSum-qwen3vl-8b consistently achieves the highest ITA-R across all length bins, and uniquely \emph{improves} from the 8k to 16k bin for most document types — a pattern not observed in any baseline — suggesting that the GRPO image-text alignment reward is especially effective when there is sufficient context for the model to identify image--text correspondences.

\begin{figure}[!htbp]
  \centering
  \includegraphics[width=0.9\columnwidth]{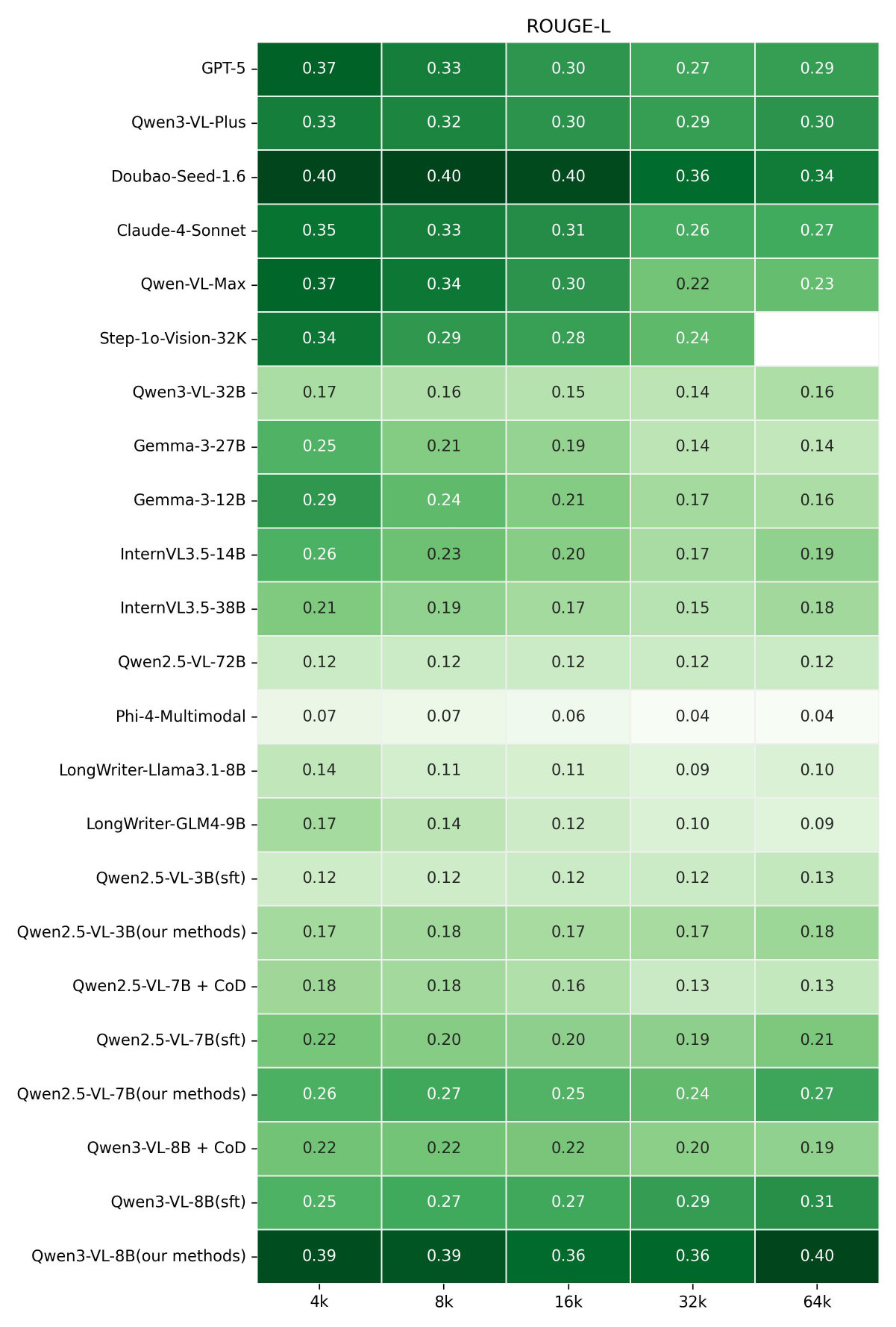}
  \caption{ROUGE-L on MMLDSum-Bench. Higher values indicate better performance.}
  \label{fig:appendix-d-rougel}
\end{figure}

\paragraph{ROUGE-L (Figure~\ref{fig:appendix-d-rougel}).}
ROUGE-L generally decreases with document length: most closed-source systems peak in the 4k--8k bins and drop toward 64k (e.g., Qwen-VL-Max $\n{0.37}\!\rightarrow\!\n{0.23}$, Claude-4-Sonnet $\n{0.35}\!\rightarrow\!\n{0.27}$, GPT-5 $\n{0.37}\!\rightarrow\!\n{0.29}$), reflecting the difficulty of preserving lexical overlap when salient evidence becomes sparser. A few models are notably flatter (Doubao-Seed-1.6: $\n{0.40}$ at 4k--16k, $\n{0.34}$ at 64k; Qwen3-VL-Plus: $\n{0.33}\!\rightarrow\!\n{0.30}$).
MMLDSum-qwen3vl-8b achieves the highest ROUGE-L in every length bin among open-source models, with a U-shaped profile ($\n{0.39}, \n{0.39}, \n{0.36}, \n{0.36}, \n{0.40}$) that is robust at both ends. The advantage is most pronounced in the 64k bin, where it ($\n{0.40}$) surpasses Doubao-Seed-1.6 ($\n{0.34}$), Qwen3-VL-Plus ($\n{0.30}$), GPT-5 ($\n{0.29}$), and Claude-4-Sonnet ($\n{0.27}$).
Compared with the SFT baseline (Qwen3-VL-8B-SFT: $\n{0.25}\!\rightarrow\!\n{0.31}$), our full two-stage model lifts ROUGE-L by $\n{0.07}$--$\n{0.14}$ across all bins, indicating that the GRPO ROUGE reward and length penalty contribute substantial gains beyond weighted SFT alone in the most challenging long-context settings.

\begin{figure*}[!t]
  \centering
  \includegraphics[width=\textwidth]{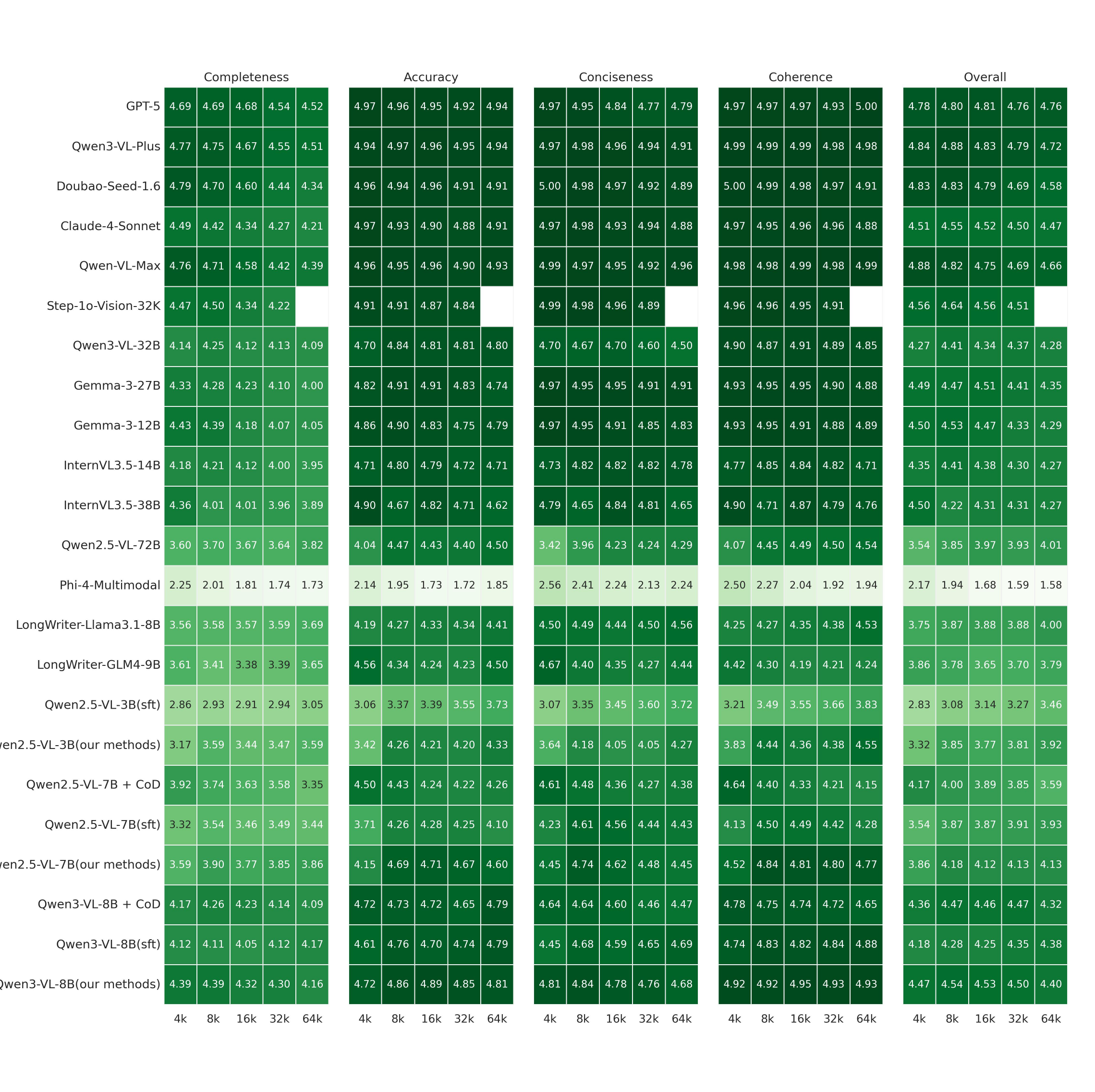}
  \caption{Heatmap visualization of GPT-4o judge-based scores on MMLDSum-Bench. Higher values indicate better performance.}
  \label{fig:appendix-d-gpt4o}
\end{figure*}

\begin{figure*}[!t]
  \centering
  \includegraphics[width=0.985\textwidth]{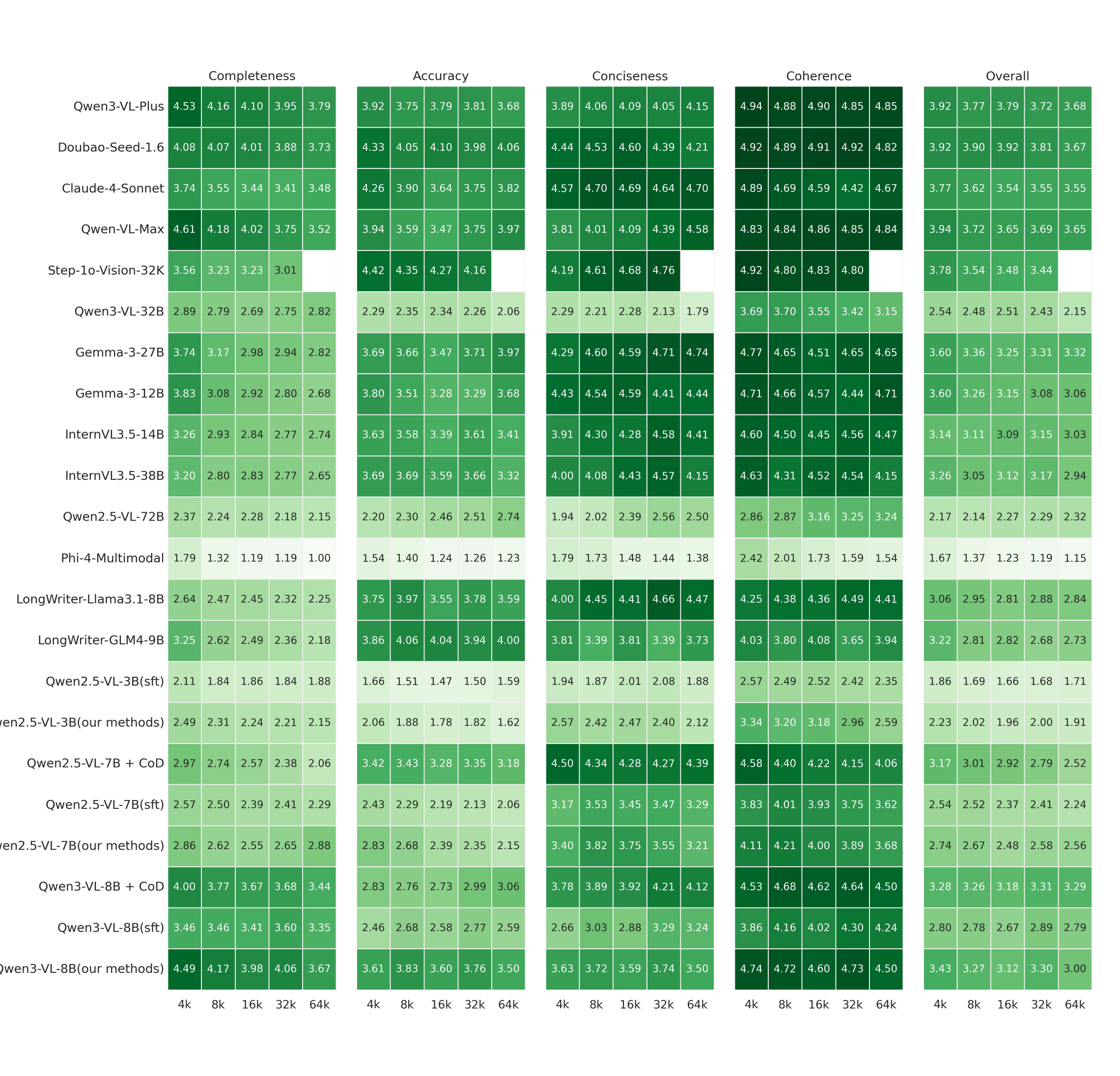}
  \caption{Heatmap visualization of GPT-5 judge-based scores on MMLDSum-Bench. Higher values indicate better performance.}
  \label{fig:appendix-d-gpt5}
\end{figure*}

\paragraph{GPT-4o and GPT-5 judge scores (Figure~\ref{fig:appendix-d-gpt4o} and Figure~\ref{fig:appendix-d-gpt5}).}
Across both judges, performance degradation with length is pronounced for most open-source models but moderate for top closed-source models and our trained models.
Closed-source models with 256k context windows (Qwen3-VL-Plus, Doubao-Seed-1.6, Qwen-VL-Max) maintain relatively stable GPT-4o scores across all five length bins, confirming that long-context ingestion capacity is a primary bottleneck for completeness.
In contrast, models with 32k context limits (Step-1o-Vision-32k, InternVL3.5-14B/38B) exhibit a clear performance drop in the 32k--64k bin; Step-1o-Vision-32k in particular shows notably depressed completeness scores in the longest bin due to forced document truncation.
GPT-5 scores generally follow the same trend as GPT-4o but with lower absolute values and wider inter-model gaps, particularly on the completeness and overall dimensions.
MMLDSum-qwen3vl-8b achieves GPT-4o overall scores competitive with Claude-4-Sonnet and Step-1o-Vision-32k across the 4k--32k range, and maintains this level into the 32k--64k bin, demonstrating that the two-stage training framework successfully extends the effective summarization range of the 8B model.

\begin{figure*}[!t]
  \centering
  \includegraphics[width=\textwidth]{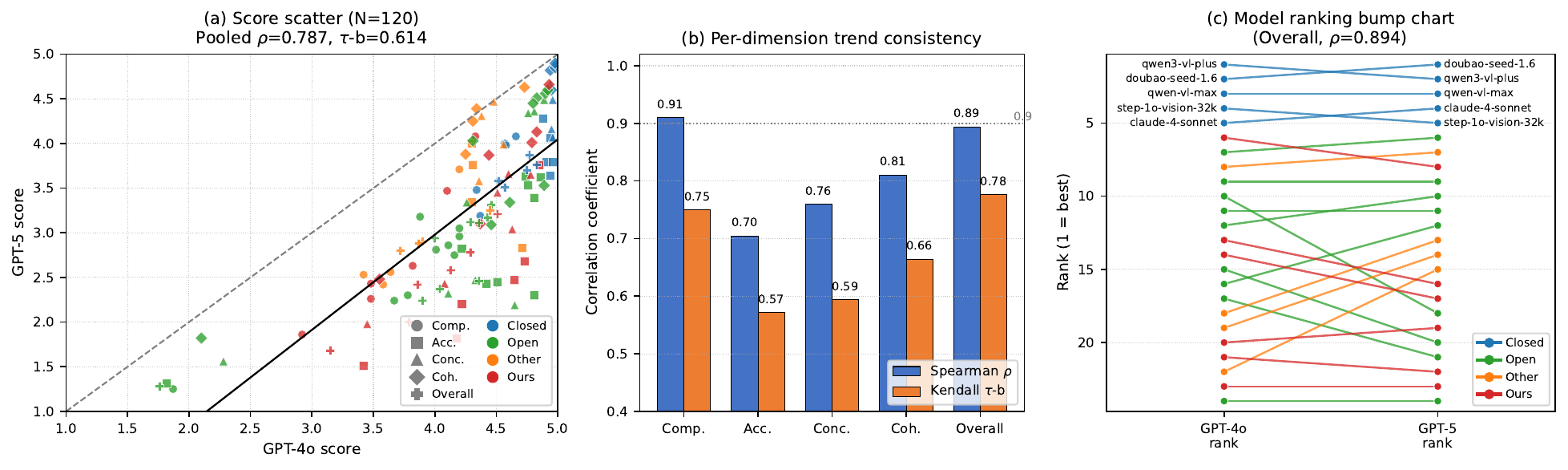}
  \caption{Quantitative cross-judge trend consistency between GPT-4o and GPT-5 on the $24$ evaluated models in Table~\ref{tab:main} (the GPT-5 self-evaluation row is excluded). \textbf{(a)}~Score scatter pooled across all five dimensions; marker shape denotes the dimension, color denotes the model category, the dashed line is $y=x$, and the solid line is the least-squares fit. \textbf{(b)}~Per-dimension Spearman~$\rho$ and Kendall~$\tau$-b; the dotted line marks the high-agreement threshold of $0.9$. \textbf{(c)}~Bump chart of model rankings: each line connects a model's GPT-4o rank to its GPT-5 rank on the Overall dimension; the top-ranked systems are nearly identical under both judges.}
  \label{fig:appendix-d-judge-agreement}
\end{figure*}

\begin{table}[!t]
\centering
\small
\setlength{\tabcolsep}{4pt}
\caption{Cross-judge trend-consistency statistics between GPT-4o and GPT-5 on the $24$ models in Table~\ref{tab:main}. Spearman~$\rho$ and Kendall~$\tau$-b measure rank agreement; Pearson~$r$ measures linear agreement. All reported correlations are significant at $p<0.001$.}
\label{tab:judge-agreement}
\begin{tabular}{lccc}
\toprule
Dimension & Spearman $\rho$ & Kendall $\tau$-b & Pearson $r$ \\
\midrule
Completeness & $0.910$ & $0.750$ & $0.890$ \\
Accuracy     & $0.704$ & $0.571$ & $0.681$ \\
Conciseness  & $0.759$ & $0.594$ & $0.771$ \\
Coherence    & $0.810$ & $0.664$ & $0.828$ \\
Overall      & $0.894$ & $0.776$ & $0.870$ \\
\midrule
\textbf{Pooled} ($N{=}120$) & $\mathbf{0.787}$ & $\mathbf{0.614}$ & $\mathbf{0.776}$ \\
\bottomrule
\end{tabular}
\end{table}

\paragraph{Cross-judge trend-consistency quantification.}
To turn the qualitative observation that GPT-4o and GPT-5 follow the same ranking into a quantitative claim, we compute three correlation coefficients between the two judges on the $24$ evaluated models in Table~\ref{tab:main} (the GPT-5 self-evaluation row is excluded). Spearman~$\rho$ and Kendall~$\tau$-b directly measure rank agreement, while Pearson~$r$ measures linear agreement; results per dimension and pooled across all five dimensions are reported in Table~\ref{tab:judge-agreement} and visualized in Figure~\ref{fig:appendix-d-judge-agreement}. All correlations are highly significant ($p<0.001$). Completeness ($\rho=0.910$) and Overall ($\rho=0.894$) exhibit the strongest agreement, indicating that the two judges essentially agree on which models are more complete and which are stronger overall; Accuracy is the least consistent ($\rho=0.704$), consistent with GPT-5 applying a stricter standard on factual claims. The bump chart in Figure~\ref{fig:appendix-d-judge-agreement}(c) further shows that the top-ranked systems are nearly identical under both judges, with only minor swaps within the top five. Together, these statistics confirm that the convergent trends reported above are not anecdotal: GPT-4o and GPT-5 produce trend-consistent rankings, supporting the validity of using both judges as complementary evaluation signals.

\paragraph{Cross-metric consistency and key takeaways.}
Aggregating the length-stratified results across the four metric families yields three consistent findings.
First, \textit{factual completeness (atomic-claim F1) and visual grounding (ITA-R) are the two metrics most sensitive to document length}, and both receive the largest absolute gains from MMLDSum-LLM's two-stage training relative to the backbone-matched SFT baseline and the CoD variant.
Second, \textit{the 32k--64k regime is the most discriminating}: systems constrained to a 32k context window (Step-1o-Vision-32K, InternVL3.5-14B/38B) degrade sharply or fail to produce outputs, whereas models with 128k+ context windows---including MMLDSum-qwen3vl-8b---retain competitive performance, confirming that adequate long-context ingestion capacity is a prerequisite for robust summarization on MMLDSum-Bench.
Third, \textit{cross-metric agreement supports the validity of the observed gains}: improvements on ITA-R and ROUGE-L (used as GRPO reward components) are accompanied by consistent improvements on the held-out evaluation signals---atomic-claim F1 and GPT-4o/GPT-5 judge scores---suggesting that the gains reflect genuine multidimensional quality improvement rather than reward fitting.

\fi
\end{document}